%% file: main.tex
\documentclass[10pt,twocolumn,letterpaper]{article}

\usepackage{cvpr}              %
\input{shortcuts}

\definecolor{cvprblue}{rgb}{0.21,0.49,0.74}
\usepackage[pagebackref,breaklinks,colorlinks,citecolor=cvprblue]{hyperref}
\hypersetup{
  colorlinks=true,
  urlcolor=ellisred, 
  linkcolor=ellisred,
  citecolor=ellisblue,
  pdfborder={0 0 0},
}

\usepackage{multirow}
\usepackage{multicol}
\usepackage{makecell}
\usepackage{xcolor}
\usepackage{algorithm} 
\usepackage{amssymb}
\usepackage{amsfonts}
\usepackage{graphicx}
\usepackage{booktabs}
\usepackage{xr}
\usepackage{epigraph}
\usepackage[T1]{fontenc}
\definecolor{mycitecolor}{rgb}{0, 0.4, 0.7}
\usepackage{comment}
\usepackage{subcaption}
\usepackage{amsmath}
\usepackage{algorithm}
\usepackage{array}
\usepackage[switch]{lineno}
\usepackage{longtable}
\usepackage{xspace}
\usepackage{url}
\usepackage{overpic}
\usepackage{ragged2e}
\usepackage{xpatch}
\usepackage{pifont}
\usepackage{enumitem}
\usepackage{bbm}
\usepackage{colortbl}
\usepackage{soul}
\usepackage{adjustbox}
\usepackage{hhline}
\usepackage{algpseudocode}

\newcommand{\cmark}{\ding{51}}%
\newcommand{\xmark}{\ding{55}}%
\usepackage{stfloats}
\usepackage{xr-hyper}
\usepackage{hyperref}
\usepackage{graphicx}
\usepackage{tikz}

\usepackage{etoolbox}
\makeatletter
\pretocmd{\chapter}{\addtocontents{toc}{\protect\addvspace{15\p@}}}{}{}
\pretocmd{\section}{\addtocontents{toc}{\protect\addvspace{5\p@}}}{}{}
\pretocmd{\subsection}{\addtocontents{toc}{\protect\addvspace{3\p@}}}{}{}
\makeatother
\usepackage{titletoc}

\def\paperID{3} %
\def\confName{CVPR}
\def\confYear{2024}

\title{
Hidden Biases of End-to-End Driving Datasets
}

\author{Julian Zimmerlin \quad \ \ \ Jens Beißwenger \quad \ \ \ Bernhard Jaeger  \quad \ \ \ Andreas Geiger  \quad \ \ \ Kashyap Chitta\\
University of Tübingen \quad \quad Tübingen AI Center\\
{\tt\footnotesize \{zim.julian, jensbeiswenger\}@gmail.com} \quad
{\tt\footnotesize \{bernhard.jaeger, a.geiger, kashyap.chitta\}@uni-tuebingen.de}}

\begin{document}

\maketitle

\input{sec_0_abstract}
\input{sec_1_intro}
\input{sec_2_prelim}
\input{sec_3_method}
\input{sec_4_experiments}

\input{sec_5_benchmarking}
\input{sec_6_conclusion}

\vspace{0.1cm}
\boldparagraph{Acknowledgements} Bernhard Jaeger and Andreas Geiger were supported by the ERC Starting Grant LEGO-3D (850533) and the DFG EXC number 2064/1 - project number 390727645. Kashyap Chitta was supported by the German Federal Ministry of Education and Research (BMBF): Tübingen AI Center, FKZ: 01IS18039A. We thank the International Max Planck Research School for Intelligent Systems (IMPRS-IS) for supporting Bernhard Jaeger and Kashyap Chitta, and Katrin Renz for the useful exchange. 

{
    \small
    \bibliographystyle{ieeenat_fullname}
    \bibliography{bibliography_long, bibliography_custom, bibliography}
}

\end{document}

%% file: shortcuts.tex
\usepackage{duckuments}
\usepackage[usenames,dvipsnames]{xcolor}

\definecolor{mediumgray}{rgb}{0.5,0.5,0.5}
\newcommand{\pmsd}[1]{{\color{mediumgray}{\scriptsize $\pm$ #1}}}

\newcommand{\figref}[1]{Fig.~\ref{#1}}
\newcommand{\secref}[1]{Section~\ref{#1}}

\newcommand{\tabref}[1]{Table~\ref{#1}}

\makeatletter
\DeclareRobustCommand\onedot{\futurelet\@let@token\@onedot}
\def\@onedot{\ifx\@let@token.\else.\null\fi\xspace}
\def\eg{e.g\onedot} 
\def\ie{i.e\onedot}

\makeatother

\newcommand{\boldparagraph}[1]{\vspace{0.1cm}\noindent{\bf #1.}}

\definecolor{darkgreen}{rgb}{0,0.7,0}
\definecolor{darkyellow}{rgb}{0.8,0.8,0}
\definecolor{bittersweet}{rgb}{1.0, 0.44, 0.37}
\definecolor{amber}{rgb}{1.0, 0.49, 0.0}
\definecolor{lgray}{rgb}{0.7,0.7,0.7}

\definecolor{color_unlabled}{rgb}{0.0,0.0,0.0}
\definecolor{color_vehicle}{rgb}{0.0,0.0,0.56}
\definecolor{color_road}{rgb}{0.5,0.25,0.5}
\definecolor{color_redlight}{rgb}{1.0,0.0,0.0}
\definecolor{color_person}{rgb}{0.859,0.078,0.234}
\definecolor{color_roadline}{rgb}{0.613,0.914,0.195}
\definecolor{color_sidewalk}{rgb}{0.953,0.137,0.906}
\definecolor{teaser_red}{RGB}{222,112,97}

\definecolor{ellisred}{rgb}{0.87,0.44,0.38} %
\definecolor{ellisgreen}{rgb}{0.69,0.90,0.52} %
\definecolor{elliscyan}{rgb}{0.29,0.77,0.74} %
\definecolor{ellisorange}{rgb}{0.89,0.55,0.28} %
\definecolor{ellisblue}{rgb}{0.41,0.61,0.86} %

\definecolor{tuedgray}{RGB}{56,55,55}
\definecolor{tuelgray}{RGB}{246,246,246}
\definecolor{tuedblue}{RGB}{26,58,91}
\definecolor{tuelblue}{RGB}{133,203,210}
\definecolor{tueoblue}{RGB}{119,221,204}
\definecolor{tueogreen}{RGB}{119,221,159}
\definecolor{tuesgreen}{RGB}{186,213,72}
\definecolor{tueyellow}{RGB}{255,221,0}
\definecolor{tuered}{RGB}{234,75,46}

\mathchardef\mhyphen="2D

%% file: sec_0_abstract.tex
\begin{abstract}
End-to-end driving systems have made rapid progress, but have so far not been applied to the challenging new CARLA Leaderboard 2.0. Further, while there is a large body of literature on end-to-end architectures and training strategies, the impact of the training dataset is often overlooked. In this work, we make a first attempt at end-to-end driving for Leaderboard 2.0. Instead of investigating architectures, we systematically analyze the training dataset, leading to new insights: (1) Expert style significantly affects downstream policy performance. (2) In complex data sets, the frames should not be weighted on the basis of simplistic criteria such as class frequencies. (3) Instead, estimating whether a frame changes the target labels compared to previous frames can reduce the size of the dataset without removing important information. By incorporating these findings, our model ranks first and second respectively on the map and sensors tracks of the 2024 CARLA Challenge, and sets a new state-of-the-art on the Bench2Drive test routes. Finally, we uncover a design flaw in the current evaluation metrics and propose a modification for future challenges. Our dataset, code, and pre-trained models are publicly available at \texttt{\url{https://github.com/autonomousvision/carla_garage}}.
\end{abstract}

%% file: sec_1_intro.tex
\section{Introduction}
\label{sec:intro}

Imitation Learning (IL) for end-to-end autonomous driving has seen great success in recent work on the CARLA simulator~\cite{Dosovitskiy2017CORL}. A key ingredient contributing to this is the scalability of IL with increased training data, which is now straightforward to collect as a result of steady progress in planning algorithms for CARLA~\cite{Chen2019CORL, Prakash2021CVPR, Jaeger2021Thesis, Zhang2021ICCV, Chitta2022PAMI, Wu2022NeurIPS, Shao2022CORL, Renz2022CORL, Jaeger2023ICCV, Ishida2024ARXIV, Li2024ECCV}. However, with the introduction of the CARLA Leaderboard 2.0, driving models now face 38 new complex scenarios. These require driving at high speeds, deviating from the center of the lane, or handling unexpected obstacles. The best planning algorithm of Leaderboard 1.0~\cite{Jaeger2023ICCV} does not solve these new scenarios, making it harder to collect the high-quality driving demonstrations needed for training IL models. As a result, there are no existing IL-based methods for Leaderboard 2.0.

In this work, we present the first attempt to tackle Leaderboard 2.0 with IL. To collect training data, we leverage the recently open-sourced PDM-Lite~\cite{Sima2024ECCV,Beißwenger2024TECH} planner, which can solve the new Leaderboard 2.0 scenarios. We then train a simple existing IL model, TransFuser++~\cite{Jaeger2023ICCV}, with minimal changes to its architecture and training objective. Instead of the model, we focus on a critical but understudied aspect of IL -- \textit{the training dataset}. In particular, the impact of factors besides the dataset scale, such as the diversity of the training distribution, is nuanced and not yet well understood. We conduct a systematic analysis of our driving dataset, leading to multiple new insights.

First, the expert's \textit{driving style}, in addition to its performance, significantly influences its suitability for IL. To develop an effective expert, it is important to base the expert's behavior on signals that are easily observable and interpretable by the IL policy, rather than relying excessively on privileged inputs. This behavior also resembles how human drivers perceive and react to their environment. Second, we find the use of frequency-based \textit{class weights}, a common approach to facilitate learning of classification tasks on imbalanced datasets, detrimental for target speed prediction in autonomous driving. Over-represented classes do not represent a single "uninteresting" mode of the data distribution-- in contrast, they may contain a mixture of both uninteresting (\eg, braking while waiting at red lights) and crucial parts of the dataset (\eg, braking for obstacles). Finally, we study \textit{data filtering} as an alternative means to assigning the importance of frames, by which we reduce our dataset size by $\sim50\%$ while maintaining performance.

Based on these findings, we train a model which safely handles urban driving in diverse scenarios to rank second in the 2024 CARLA challenge and first on the Bench2Drive test routes~\cite{Jia2024NEURIPS}. We then theoretically demonstrate how the performance metrics used by the leaderboard inadvertently encourage participants to terminate evaluation routes prematurely, and propose changes to the metrics that can solve this problem for future challenges. An extended report of all our experiments and findings is available \href{https://kashyap7x.github.io/assets/pdf/students/Zimmerlin2024.pdf}{at this link}.

%% file: sec_2_prelim.tex
\section{Preliminaries}
\label{sec:prelim}

In this section, we provide an overview of our task and baselines. The task involves urban navigation along routes with complex scenarios. Each route is a list of GNSS coordinates called target points (TPs) which can be up to 200 m apart.

\boldparagraph{Metrics} For the following experiments, we use the official CARLA closed-loop metrics. Our main metric is the Driving Score (DS) which multiplies Route Completion (RC) with the Infraction Score (IS). RC is the percentage of the route completed. IS is a penalty factor, starting at 1.0, which gets reduced multiplicatively with each infraction.

\input{fig/fig_scenario_distribution}

\boldparagraph{Benchmark} To train and evaluate agents, {Leaderboard 2.0} provides 90 training routes on Town12 and 20 {validation} routes on Town13 which on average are 8.67 km and 12.39 km long respectively. Each route contains around 100 scenarios, distributed as shown in Figure \ref{fig:scenario_distribution}. We split them into short routes, each containing only a single scenario.  This allows for more accurate performance evaluation per scenario type. After splitting, we sample up to 15 routes per scenario type without replacement to create the {Town13 short} benchmark. There are 38 scenario types, but in some cases, fewer (or no) routes are available, which gives a total of 400 routes from 36 scenarios in this benchmark. As the calculation of \textit{MinSpeedInfractions} is unsuited to short routes, we exclude them from the IS metric on Town13 short.

\boldparagraph{Training dataset}  We reproduce TransFuser++~\cite{Jaeger2023ICCV} on our benchmark using data collected with the PDM-Lite expert \cite{Sima2024ECCV,Beißwenger2024TECH}. We choose PDM-Lite as it achieves state-of-the-art DS on the official validation routes. Unlike other concurrent Leaderboard 2.0 planners \cite{Zhang2024ARXIV, Li2024ARXIV}, it is also publicly available and possible to modify. We sample from the shortened training routes with replacement to obtain a set of 50 routes per scenario, on which we collect a training dataset using our expert (198k frames). The dataset contains RGB images with a resolution of 384x1024 pixels, LiDAR point clouds, and the training labels needed for TF++ (path checkpoints, expert target speed, and auxiliary labels such as BEV semantics and vehicle/pedestrian bounding box predictions). Additionally, we collect data on Towns 01-05 and 10, which contain the six scenarios from Leaderboard 1.0 (139k frames), for a total of 337k frames. For the final leaderboard submissions, we also include training data from the provided validation routes on Town13 (50 short routes per scenario), adding 194k frames (531k in total).

\begin{figure}[t]
    \centering
    \includegraphics[width=\linewidth]{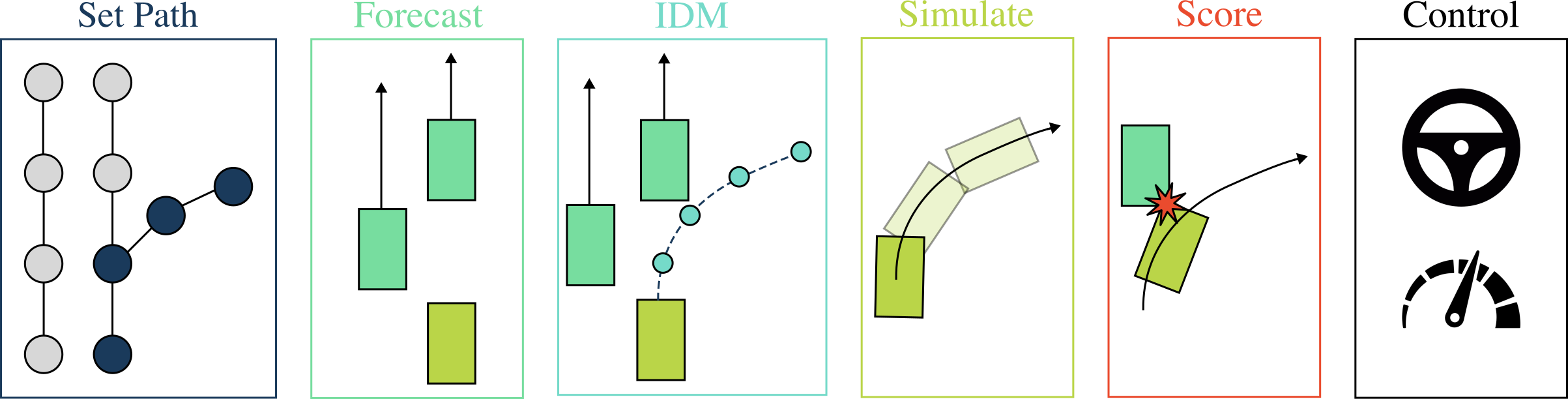}
    \caption{\textbf{PDM-Lite}~\cite{Sima2024ECCV,Beißwenger2024TECH}. This open-source rule-based planner solves all 38 scenarios of CARLA Leaderboard 2.0.}
    \label{fig:carla_pdm_lite}
    \vspace{-0.4cm}
\end{figure}

\noindent \textbf{PDM-Lite}~\cite{Sima2024ECCV,Beißwenger2024TECH} is a rule-based approach for collecting data in Leaderboard 2.0. Inspired by PDM-Closed~\cite{Dauner2023CORL}, it consists of six stages (\figref{fig:carla_pdm_lite}):

\begin{itemize}
    \item First, it creates a dense path of spatially equidistant points using the A* planning algorithm, given sparse TPs from the simulator. For new scenarios that require leaving this path, a short segment of the route where the scenario will be spawned is shifted laterally towards an adjacent lane.
    \item It forecasts dynamic agents for 2s into the future, assuming that they maintain their previous controls.
    \item It selects a leading actor and generates a target speed proposal using the Intelligent Driver Model~\cite{Treiber2000PRE}.
    \item The target speed proposal is converted into an actual expected sequence of ego-vehicle bounding boxes in closed-loop by using a kinematic bicycle model.
    \item Having forecasted all actors, it checks for bounding box intersections between the simulated ego vehicle and other vehicles. It scores the ego vehicle's motion accordingly: if it detects an intersection, it rejects the IDM target speed proposal, and sets the target speed to zero.
    \item The steering value is estimated with a lateral PID controller, which minimizes the angle to a selected point along the path ahead. For the throttle and brake predictions, it employs a linear regression model using features extracted based on the current speed and target speed.
\end{itemize}

\begin{figure}[t]
\begin{center}
   \includegraphics[width=\linewidth]{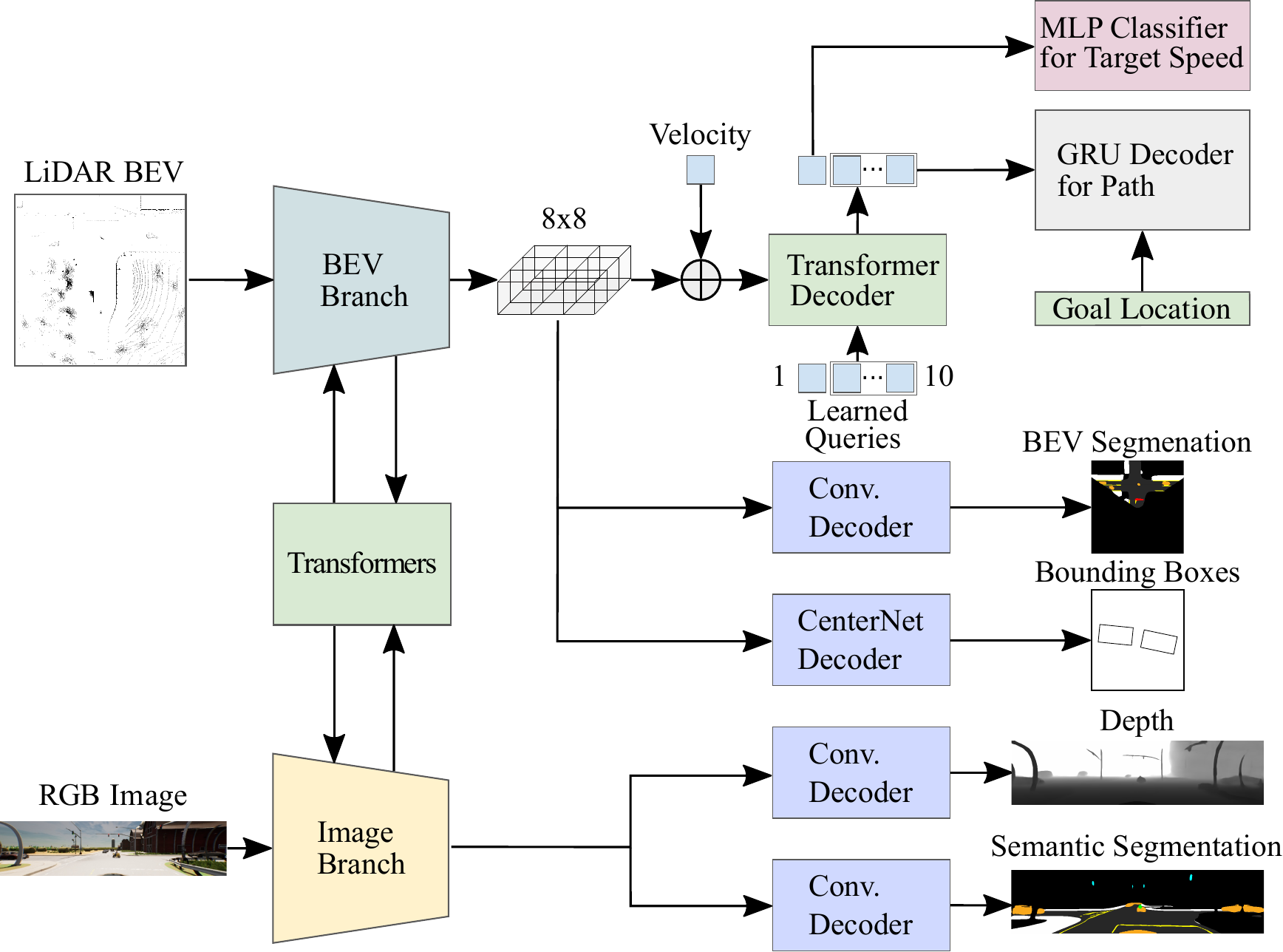}
\end{center}
\vspace{-0.3cm}
\caption{\textbf{TF++}~\cite{Jaeger2023ICCV}. This end-to-end imitation learning approach is the best publicly available baseline for CARLA.}
\label{fig:architecture}
\vspace{-0.5cm}
\end{figure}

\noindent \textbf{TransFuser++}~\cite{Jaeger2023ICCV} is the best-performing open-source model on Leaderboard 1.0 (\figref{fig:architecture}). Given sensor inputs, it predicts a target speed and desired path which are input to a controller module to drive the vehicle. We require two changes compared to~\cite{Jaeger2023ICCV} for compatibility with PDM-Lite:

\begin{itemize}
    \item \textbf{Two-hot labels.} While the rule-based planner in \cite{Jaeger2023ICCV} uses only 4 different target speed classes up to 8m/s (28.8km/h), PDM-Lite operates with a continuous range of target speed values up to 20m/s (72km/h). To solve target speed regression with a classification module, we employ \textit{two-hot labels}~\cite{Farebrother2024ARXIV}. This method converts a continuous value into a two-hot representation by interpolating between one-hot labels of the two nearest classes. For instance, with our 8 speed classes ([0.0, 4.0, 8.0, 10, 13.89, 16, 17.78, 20] m/s), a target speed of 3.0m/s is represented as $[0.25, 0.75, 0, 0, 0, 0, 0, 0]$. These speed classes were selected by analyzing the distribution of target speeds chosen by PDM-Lite in our dataset.
    \item \textbf{Dynamic lookahead controller.} For stable lateral control at the high speeds required by Leaderboard 2.0, it is advantageous to adjust the distance of the point selected to follow along the ego vehicle's predicted path based on the current speed. TF++ predicts a set of 10 checkpoints, each spaced 1m apart, with the first checkpoint located 2.5 meters from the vehicle center. The distance of the checkpoint to which the lateral controller minimizes the angle is determined by the formula $d = (0.098v+0.192)$, where $v$ is the ego's speed in km/h. We round down to the nearest available predicted checkpoint. This scaling ensures that at low speeds, the controller selects a closer point, facilitating tight turns, while at high speeds, it selects a distant point, resulting in more stable steering.
\end{itemize}

\boldparagraph{Implementation} We use a cosine annealing learning rate schedule~\cite{loshchilov2017iclr} with $lr_0 = 3\cdot10^{-4}, T_0 = 1, T_{mult}=2$ and train our models for 31 epochs. We train each model on four A100 GPUs with a total batch size of 64, which takes 2-3 days depending on the architecture. 

%% file: fig/fig_scenario_distribution.tex
\begin{figure}[t]
\begin{center}
   \includegraphics[width=\linewidth]{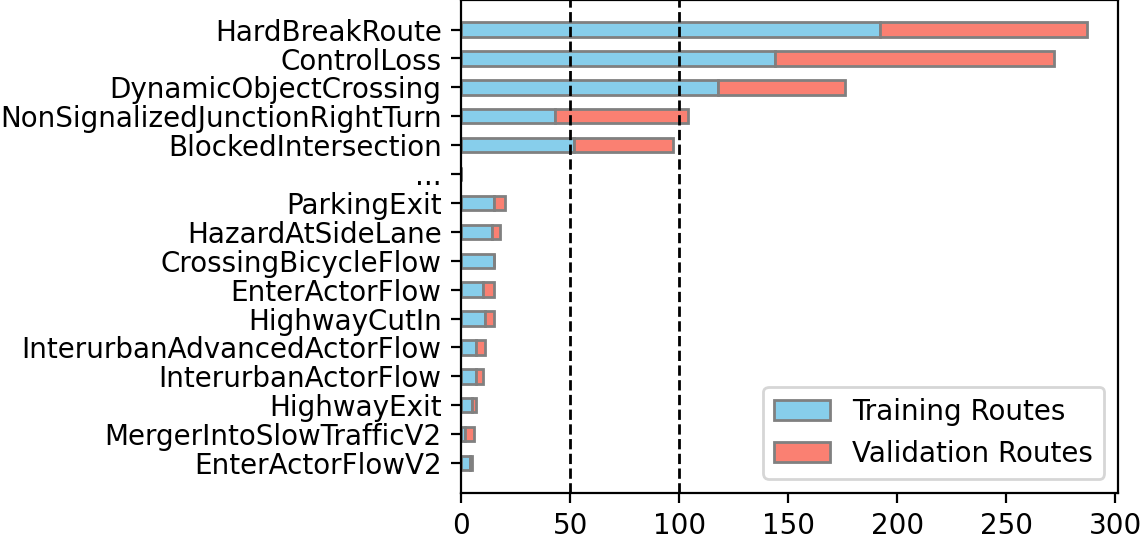}
\end{center}
\vspace{-0.6cm}
\caption{\textbf{Scenario distribution} in the available long routes.}
\label{fig:scenario_distribution}
\vspace{-0.3cm}
\end{figure}

%% file: sec_3_method.tex
\section{Hidden Biases}
\label{sec:method}

In this section, we present the main findings of our study. 

\boldparagraph{Expert style} While expert \textit{performance} is often reported in prior work, the manner in which it achieves that performance, \ie, expert \textit{style}, is often overlooked. Although harder to quantify, it is an important aspect to consider for IL. For instance, consider PDM-Lite's behavior when approaching pedestrians (\figref{fig:expert_style}). By default, the expert slows down when a pedestrian will enter the driving path, even if the pedestrian is still obstructed by a parked vehicle and not clearly visible. We make minor adjustments to the IDM parameters, with which the expert brakes rather sharply, coming to a stop at a distance of roughly 4 meters in front of the more visible pedestrian. This leads to a $\sim4\times$ decrease in pedestrian collisions for models trained on the adjusted expert data. Notably, this update does not affect the expert's own pedestrian collision rate. The improvement is likely due to the adjusted behavior providing a clear braking signal for the model to learn from (a pedestrian visible directly in front of the ego vehicle). The default behavior requires the model to generalize across various situations where a pedestrian \textit{might} appear further ahead.

\input{fig/fig_expert_style}

\boldparagraph{Target speed weights} Training TF++ involves using class weights in a target speed classification loss, which are calculated anti-proportionally to the number of occurrences of the respective class in the dataset~\cite{Jaeger2023ICCV}. This means that classes that appear frequently get a lower weight than those that appear rarely. We find that removing these weights significantly improves the performance of TF++ on our task (Table \ref{tab:speedweights}). We believe this is due to the weight of class 0 (braking), the most common in the dataset. While some part of the data for class 0 is redundant (e.g., waiting at red lights), some frames are among the most crucial for avoiding infractions, such as coming to a stop in front of stop signs or pedestrians. With a low weight on the target speed loss for these frames, ignoring short braking phases in these situations is an easy shortcut for the model to fall into.
\input{tab/tab_speedweights}

\boldparagraph{Data filtering} As an alternative to measure importance of frames, we propose the use of heuristics that estimate whether a frame changes the model's target labels compared to previous frames. More precisely, we keep all frames that change the target speed by more than 0.1m/s, or the angle to any of the predicted path checkpoints by more than 0.5° compared to the previous frame ($\sim$40\% of all frames). Additionally, we randomly retain 14\% of the remaining frames and discard the rest, for a total filtered dataset containing 51\% of all available frames. We then train with $2\times$ the number of epochs to keep the total number of gradient updates similar. In Table \ref{tab:leaderboard}, we present the results of our proposed filtering strategy on the official Leaderboard. By reducing the dataset size by 49\%, with slightly improved performance, we show that our heuristic effectively removes redundant frames without losing information. 

\input{tab/tab_leaderboard}

%% file: fig/fig_expert_style.tex
\begin{figure}[t!]
        \begin{tikzpicture}
        \node[anchor=south west,inner sep=0] (image) at (0,0) {\includegraphics[width=\linewidth]{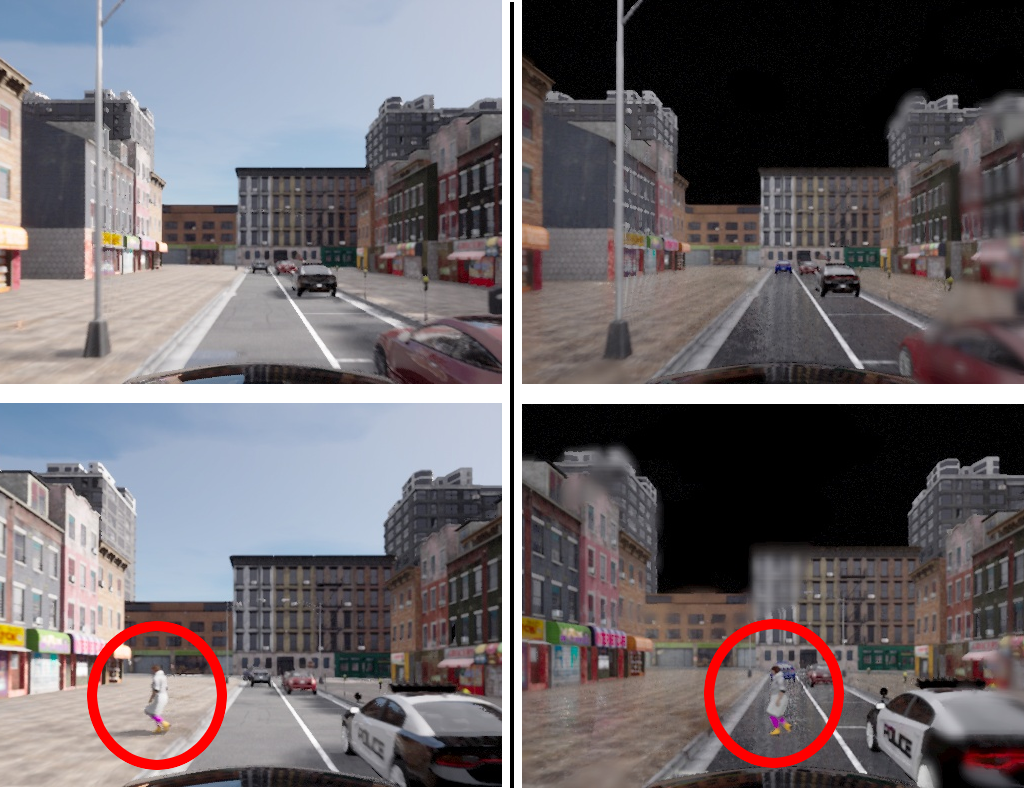}};

        \begin{scope}[x={(image.south east)},y={(image.north west)}]
            \node[anchor=north west, text=white] at (0.02,0.98) {%
                \parbox{0.47\linewidth}{\raggedright\small
                    Default, target spd: 4.9m/s
                }
            };

            \node[anchor=north west, text=white] at (0.02,0.48) {%
                \parbox{0.47\linewidth}{\raggedright\small
                    Default, target spd: 8.3m/s
                }
            };

            \node[anchor=north west, text=white] at (0.52,0.98) {%
                \parbox{0.47\linewidth}{\raggedright\small
                    Adjusted, target spd: 9.5m/s
                }
            };

            \node[anchor=north west, text=white] at (0.52,0.48) {%
                \parbox{0.47\linewidth}{\raggedright\small
                    Adjusted, target spd: 3.6m/s
                }
            };
        \end{scope}

    \end{tikzpicture}
    \vspace{-0.5cm}
    \caption{\textbf{Expert style compared on the same route.} The default PDM-Lite brakes early (left), when the pedestrian is hardly visible in the image, while the adjusted expert brakes later (right).}
    \label{fig:expert_style}
    \vspace{-0.5cm}
\end{figure}
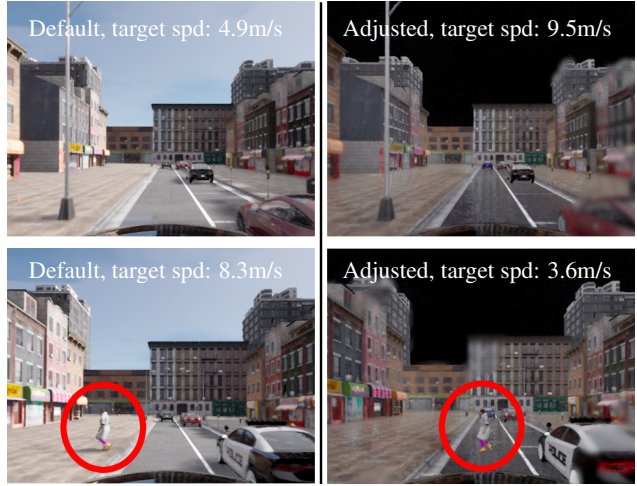

%% file: tab/tab_speedweights.tex
\begin{table}[h!]
\small
    \centering
    \begin{tabular}{c | c c c}
        \toprule
        \textbf{Speed Weights} & \textbf{DS}$\uparrow$ & \textbf{RC} $\uparrow$ & \textbf{IS} $\uparrow$ \\
        \midrule
        \cmark & 82 \pmsd{1} & 98 \pmsd{0} & 0.83 \pmsd{0.01}\\ %
        \xmark & 85 \pmsd{0} & 99 \pmsd{0} & 0.85 \pmsd{0.00} \\
        \bottomrule
    \end{tabular}
    \caption{\textbf{Speed weights.} Results on Town13 short, reported over 3 training seeds. Weights: [0.29, 1.30, 0.69, 0.81, 4.43, 4.76, 3.90, 2.41] for the speeds [0.0, 4.0, 8.0, 10, 13.89, 16, 17.78, 20] m/s.}
    \label{tab:speedweights}
    \vspace{-0.3cm}
\end{table}

%% file: tab/tab_leaderboard.tex
\begin{table}[t!]
\small
\centering
    \begin{tabular}{l| c c c}%
        \toprule
        \textbf{Model} & \textbf{DS} $\uparrow$ & \textbf{RC} $\uparrow$ & \textbf{IS} $\uparrow$  \\
        \midrule
        LRM~\cite{Rosero2024Sensors} & 1.2 & 9.6 & 0.31  \\
        Kyber-E2E~\cite{Zhang2024ARXIV} & 3.5 & 8.5 & 0.50  \\
        CarLLaVA~\cite{Renz2024ARXIV} & 6.9 & 18.1 & 0.42\\
        \midrule
        TF++ (no filtering) & 5.2 & 11.3 & 0.48  \\
        TF++ (w/ filtering) & 5.6 & 11.8 & 0.47  \\
        \bottomrule
    \end{tabular}
    \caption{\textbf{Filtering improves scores on CARLA Leaderboard 2.0.} Secret test routes (Town 14). TF++ (Ours) outperforms prior modular pipelines~\cite{Rosero2024Sensors, Zhang2024ARXIV}, and places 2nd overall.}
    \label{tab:leaderboard}
    \vspace{-0.3cm}
\end{table} 	 	

%% file: sec_4_experiments.tex
\section{Additional Insights}
\label{sec:experiments}

In this section, we analyze the behavior of our models. In Table \ref{tab:main_experiments}, we present additional experimental results with no speed weights and no filtering. Models marked with:
\begin{itemize}
    \item "\textit{Big}" use the default regnety\_032~\cite{Radosavovic2020arxiv} architecture of TF++ for the image and LiDAR perception modules instead of ResNet34 used in our "Base" setting.
    \item "\textit{Pre}" use two-stage training, where we first pre-train exclusively with perception losses (BEV semantics, bounding boxes, image depth, image semantics) for 15 epochs, before training for 31 epochs with all losses.
    \item "\textit{Ens}" are ensemble models, averaging predictions from 3 models trained with different random seeds.
\end{itemize}

Ensembling ("Ens") and two-stage training ("Pre") provide small improvements. To react better to vehicles that are further away, we experiment with extending the LiDAR range in front of the ego to 64m from 32m ("L64m"). This also results in a small improvement, albeit at the cost of increased training time. Giving the next two target points as input instead of only one ("2TPs") fails to increase performance. Our final models (used in \tabref{tab:leaderboard} and \secref{sec:benchmarking}) combine the benefits of "Big", "Pre", and "Ens".

\input{tab/tab_main_experiments}
\input{fig/Scen_COTW}

\boldparagraph{Failures} We now discuss scenarios where the "Base" TF++ model fails often. We visualize the camera image input and a bird's-eye-view (BEV) image showing observed LiDAR hits, the model's path predictions, the target points used as inputs, and the auxiliary BEV semantics predictions. We use the following colors:

\begin{center}
\begin{tabular}{|c|l|}
\hline
\tikz\draw[blue, fill=blue] (0,0) circle (.5ex); & \textbf{Blue}: Path predictions \\ \hline
\tikz\draw[red, fill=red] (0,0) circle (.5ex); & \textbf{Red}: Target point (used as model input) \\ \hline
\cellcolor{gray!30} & \textbf{Grey}: Road (semantics) \\ \hline
\cellcolor{yellow} & \textbf{Yellow}: Road marking (semantics)\\ \hline
\cellcolor{green!30} & \textbf{Light Green}: Green traffic light (semantics) \\ \hline
\cellcolor{orange!60} & \textbf{Light Orange}: Vehicle (semantics)
\\ \hline
\cellcolor{green} & \textbf{Green}: Ego vehicle or pedestrian (bounding box) \\ \hline
\cellcolor{orange} & \textbf{Orange}: Vehicle (bounding box) \\ \hline
\end{tabular}
\end{center}

\boldparagraph{ConstructionObstacleTwoWays} In this scenario, the ego vehicle must pass an obstacle by moving into an adjacent lane with oncoming traffic. Figure~\ref{fig:Scen_COTW} illustrates a common failure mode, where TF++ fails to merge back to its original lane. At first, the model successfully waits for a sufficiently large gap in the oncoming traffic and switches to the adjacent lane. As long as the traffic cones marking the construction site are visible in the camera image, its path predictions correctly indicate a lane change back to the original lane. However, as shown in the final frame, once the traffic cones disappear from the camera image, the model erroneously predicts staying in the left lane. Notably, the traffic cones are still visible in the LiDAR, suggesting an over-reliance on the camera for this scenario. After staying in the wrong lane, the ego vehicle often collides with oncoming traffic.

\boldparagraph{SignalizedJunctionLeftTurn} Figure~\ref{fig:Scen_SJLT} shows a common failure case where the model does not react adequately to another actor, leading to a vehicle collision. After the traffic light turns green, the vehicle accelerates and turns without reacting to the oncoming vehicle in the lane it needs to cross. After the collision, the model is able to recover, returning to its lane and completing the route.

\boldparagraph{VehicleTurningRoutePedestrian} In this scenario, the model must make an unproteced turn through dense traffic and encounters a pedestrian on the road during or directly after the turn. Figure~\ref{fig:Scen_VTRP} shows two corresponding failure cases. In the first case (top), the model fails to execute the turn through very dense traffic, where the margins for selecting the right moment to accelerate are extremely narrow. After colliding with a vehicle, the model also collides with a pedestrian that walks into the ego vehicle while it is stationary. The second example (bottom) depicts an instance of this scenario at night, where TF++ does not collide with another vehicle, but fails to recognize the pedestrian hazard in time. Note that the pedestrian is barely visible until illuminated by the ego vehicle's headlights, which is only a couple of frames before the collision. This failure is likely due to covariate shift, since the expert would brake earlier in this situation, even before the pedestrian becomes visible in the RGB image. By the time the pedestrian is revealed by the headlights, the model is already outside its training distribution, where it has not learned a braking reflex.

\input{fig/Scen_SJLT}

\input{fig/Scen_VTRP}

\boldparagraph{YieldToEmergencyVehicle} In this scenario, the ego vehicle must yield to an emergency vehicle approaching from behind on a multi-lane highway. TF++ fails here since it is impossible to distinguish emergency vehicles from regular vehicles using the LiDAR alone. Thus, solving this scenario requires the inclusion of a back camera.

%% file: tab/tab_main_experiments.tex
\begin{table}[t!]
\small
\centering
    \begin{tabular}{c| c c c}%
        \toprule
        \textbf{Setting} & \textbf{DS} $\uparrow$ & \textbf{RC} $\uparrow$ & \textbf{IS} $\uparrow$ \\
        \midrule
        Base & 85 \pmsd{0} & 99 \pmsd{0} & 0.86 \pmsd{0.00} \\
        Big & {85} \pmsd {1} & {98} \pmsd {1} & {0.86} \pmsd {0.01} \\
        Pre & {87} \pmsd {1} & {98} \pmsd {0} & {0.87} \pmsd {0.01} \\
        Ens & 86 & 98 & 0.87 \\
        L64m & 86 & 97 & 0.87 \\
        2TPs & 82 & 96 & 0.85 \\
        \midrule
        \textit{PDM-Lite}~\cite{Sima2024ECCV} & \textit{99} & \textit{100} & \textit{0.99}  \\
        \bottomrule
    \end{tabular}
    \caption{\textbf{Ablations on Town13 short.} Std over 3 training seeds where available. Training on 337k frames (excluding Town13).}
    \label{tab:main_experiments}
    \vspace{-0.5cm}
\end{table}

%% file: fig/Scen_COTW.tex
\begin{figure*}
\begin{center}
    \includegraphics[width=0.328\textwidth, clip=true, trim=150pt 300pt 150pt 0pt]{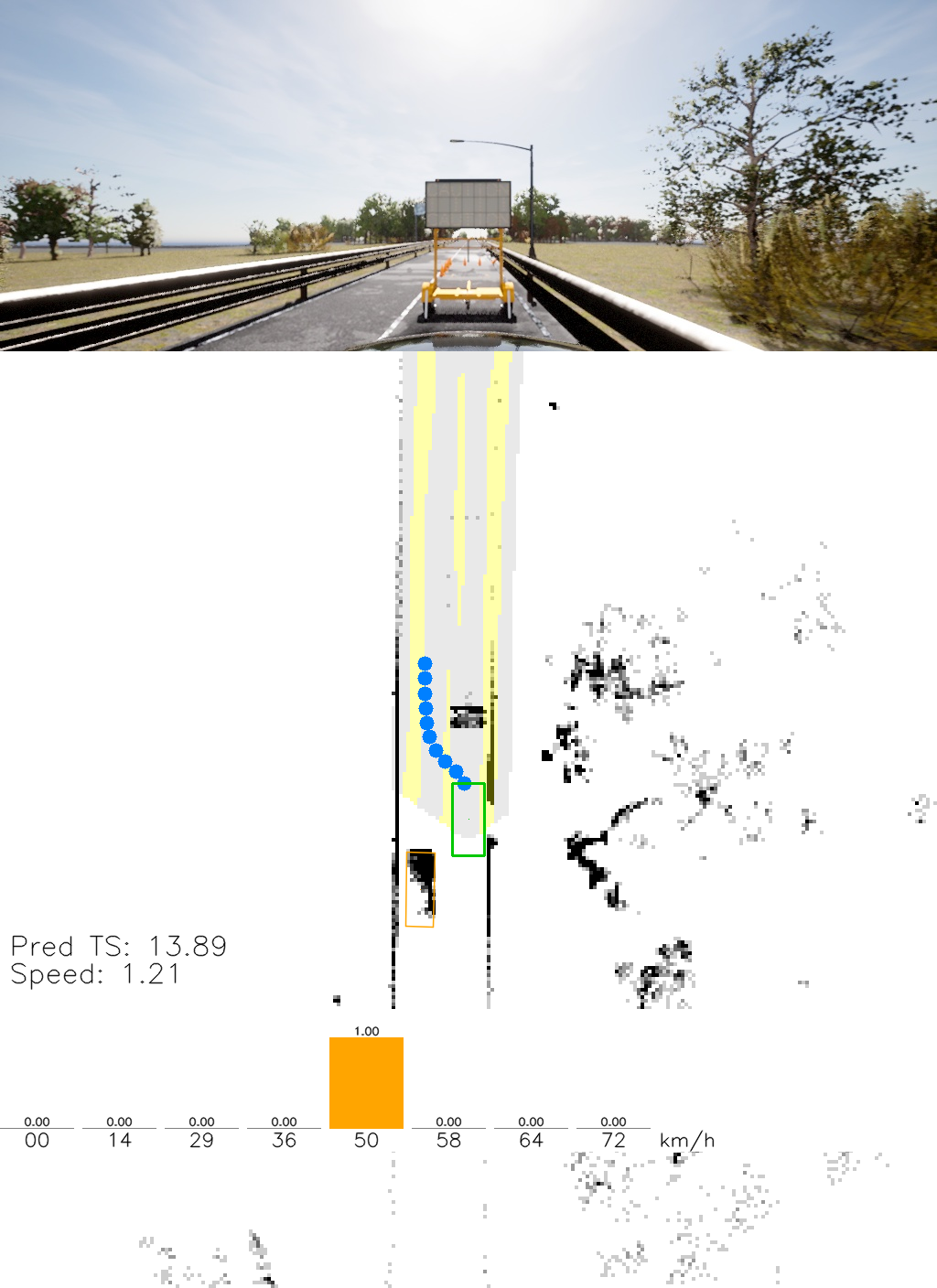}
    \includegraphics[width=0.328\textwidth, clip=true, trim=150pt 300pt 150pt 0pt]{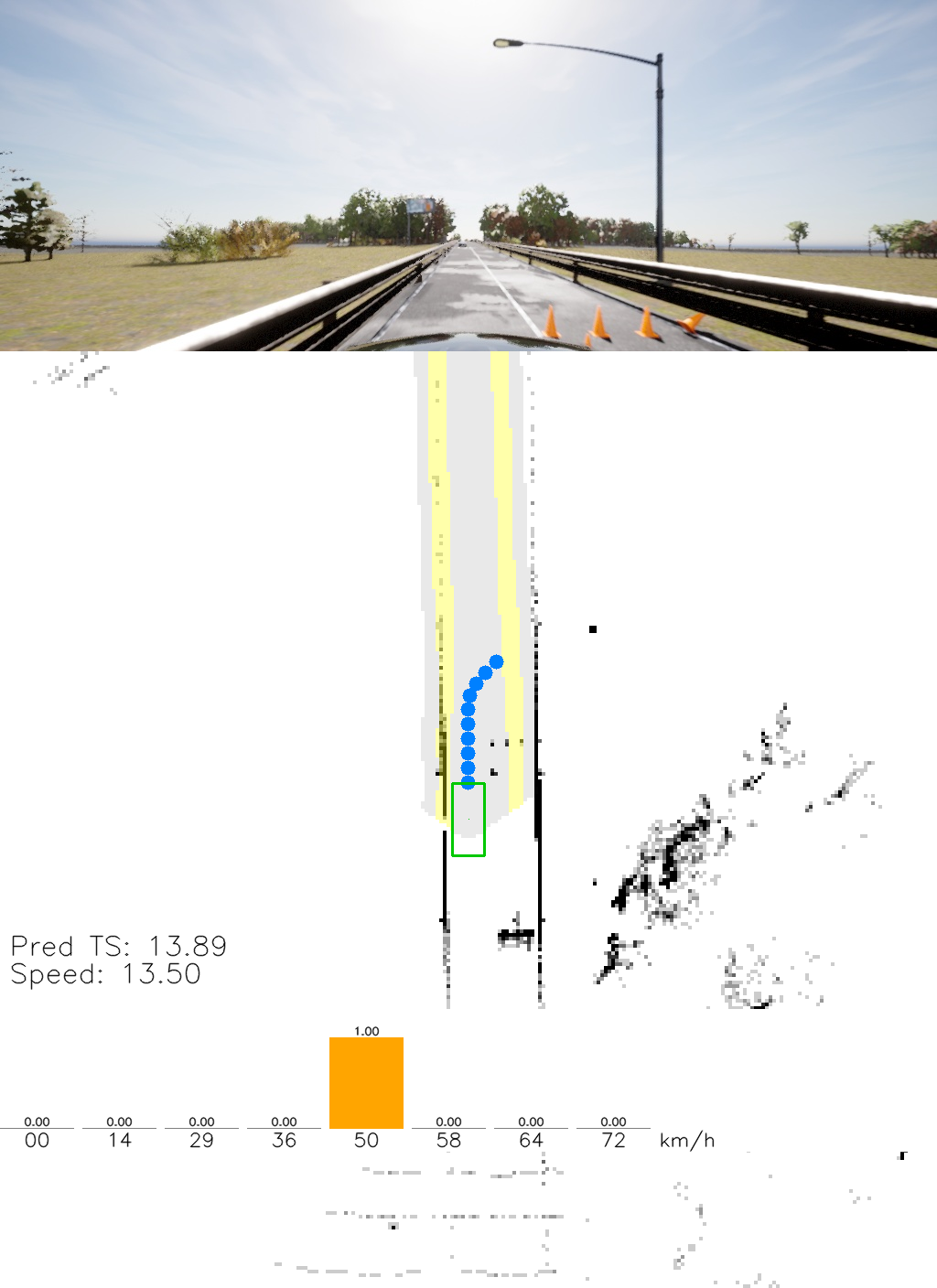}
    \includegraphics[width=0.328\textwidth, clip=true, trim=150pt 300pt 150pt 0pt]{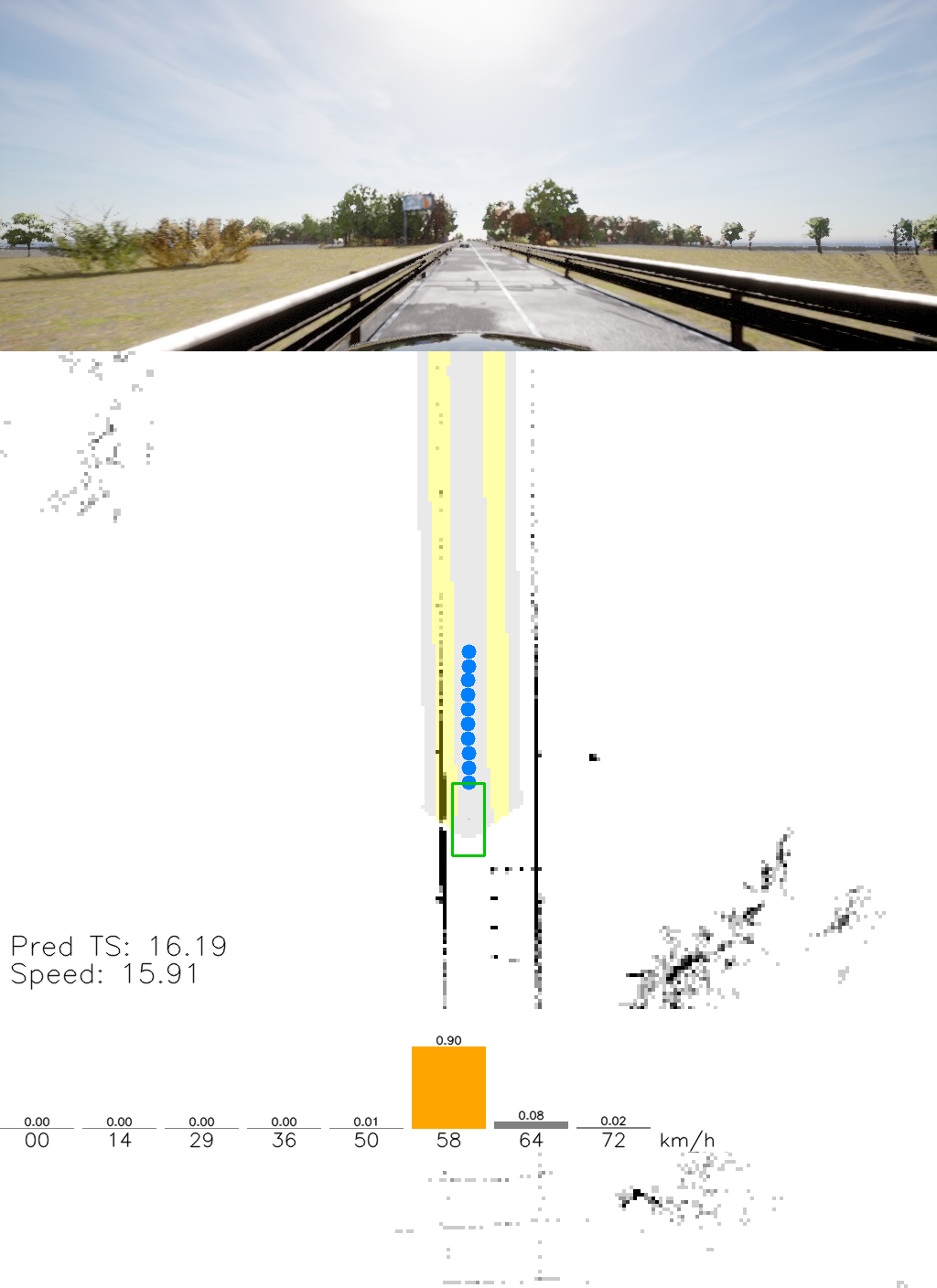}
\end{center}
\vspace{-0.5cm}
\caption{\textbf{ConstructionObstacleTwoWays.} When traffic cones are not visible any more, TF++ forgets to return to its original lane.}
\label{fig:Scen_COTW}
\vspace{-0.3cm}
\end{figure*}

%% file: fig/Scen_SJLT.tex
\begin{figure*}
\begin{center}
    \includegraphics[width=0.328\textwidth, clip=true, trim=150pt 300pt 150pt 0pt]{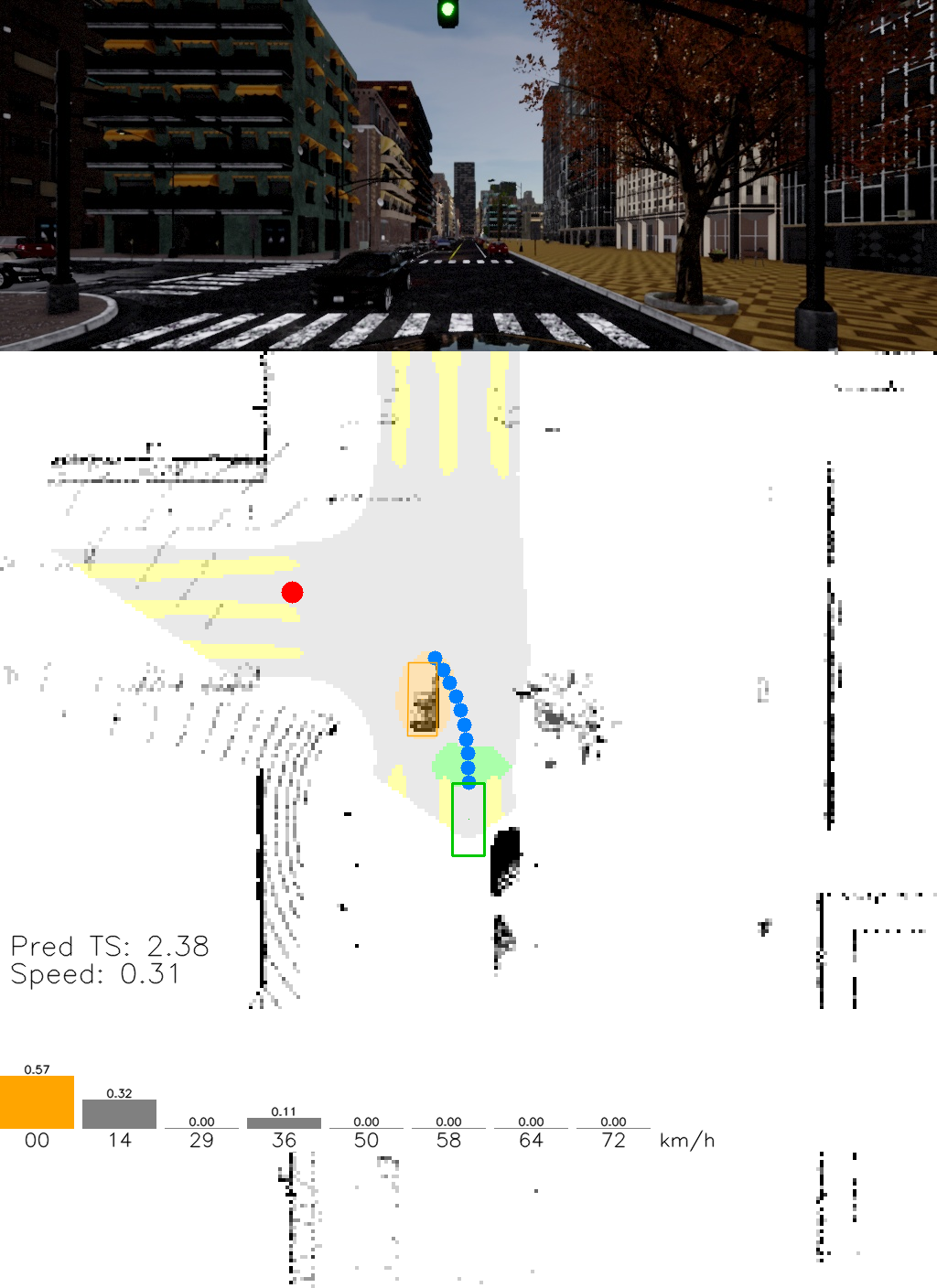}
    \includegraphics[width=0.328\textwidth, clip=true, trim=150pt 300pt 150pt 0pt]{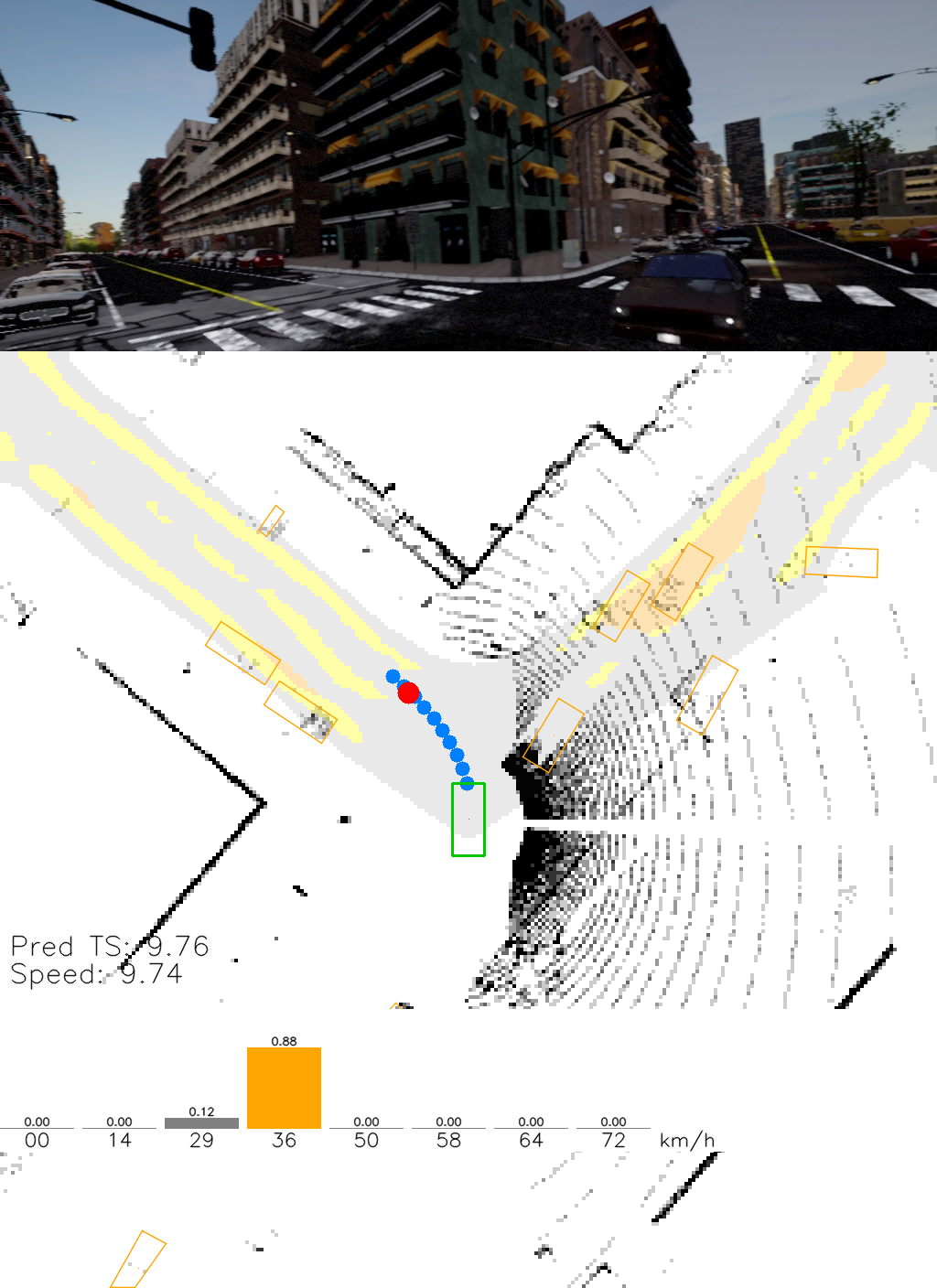}
    \includegraphics[width=0.328\textwidth, clip=true, trim=150pt 300pt 150pt 0pt]{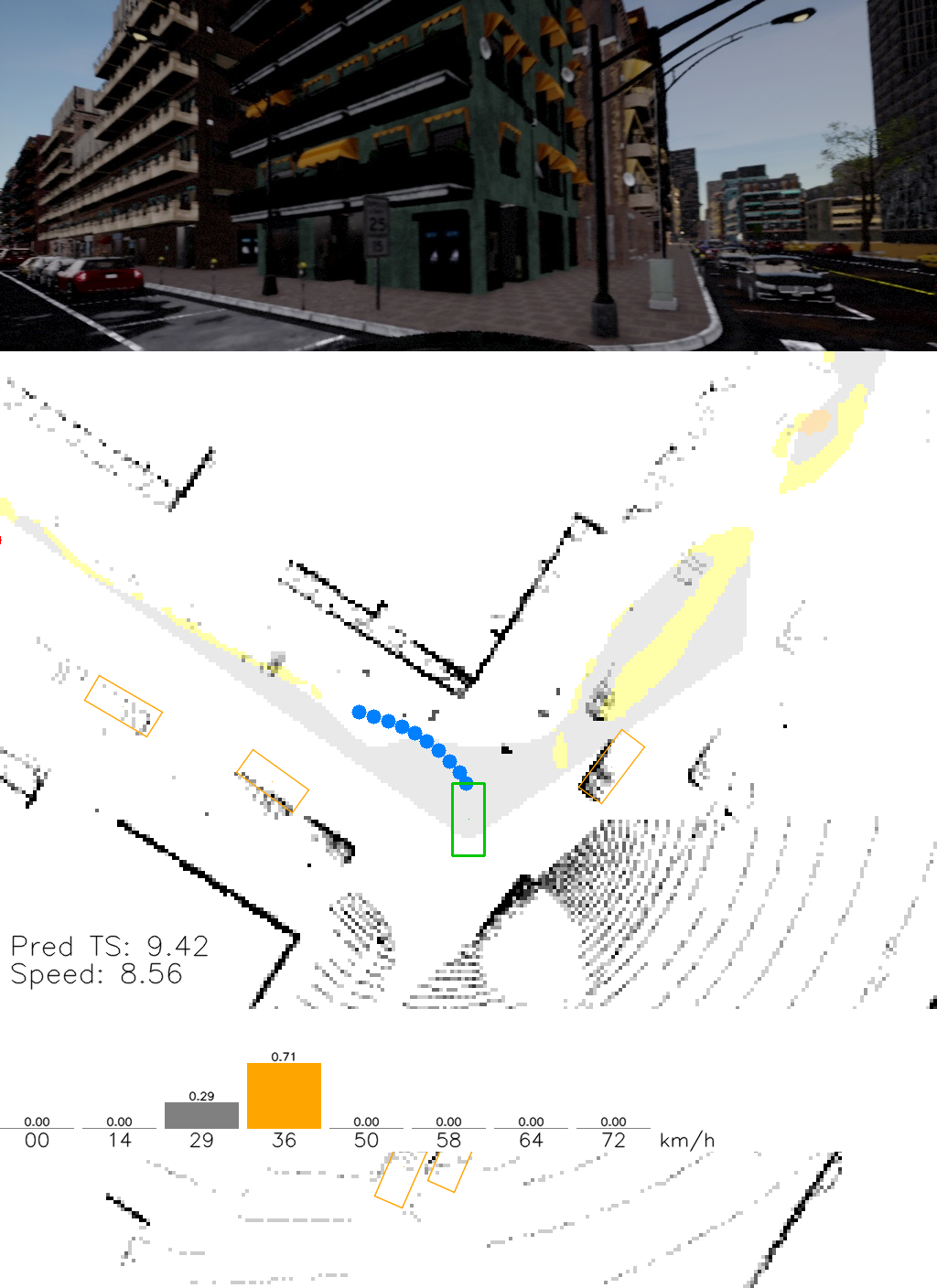}
\end{center}
\vspace{-0.5cm}
\caption{\textbf{SignalizedJunctionLeftTurn.} In a left turn after waiting at traffic light, oncoming traffic is neglected, leading to vehicle collision.}
\label{fig:Scen_SJLT}
\end{figure*}

%% file: fig/Scen_VTRP.tex
\begin{figure*}
\begin{center}
    \includegraphics[width=0.328\textwidth, clip=true, trim=150pt 300pt 150pt 0pt]{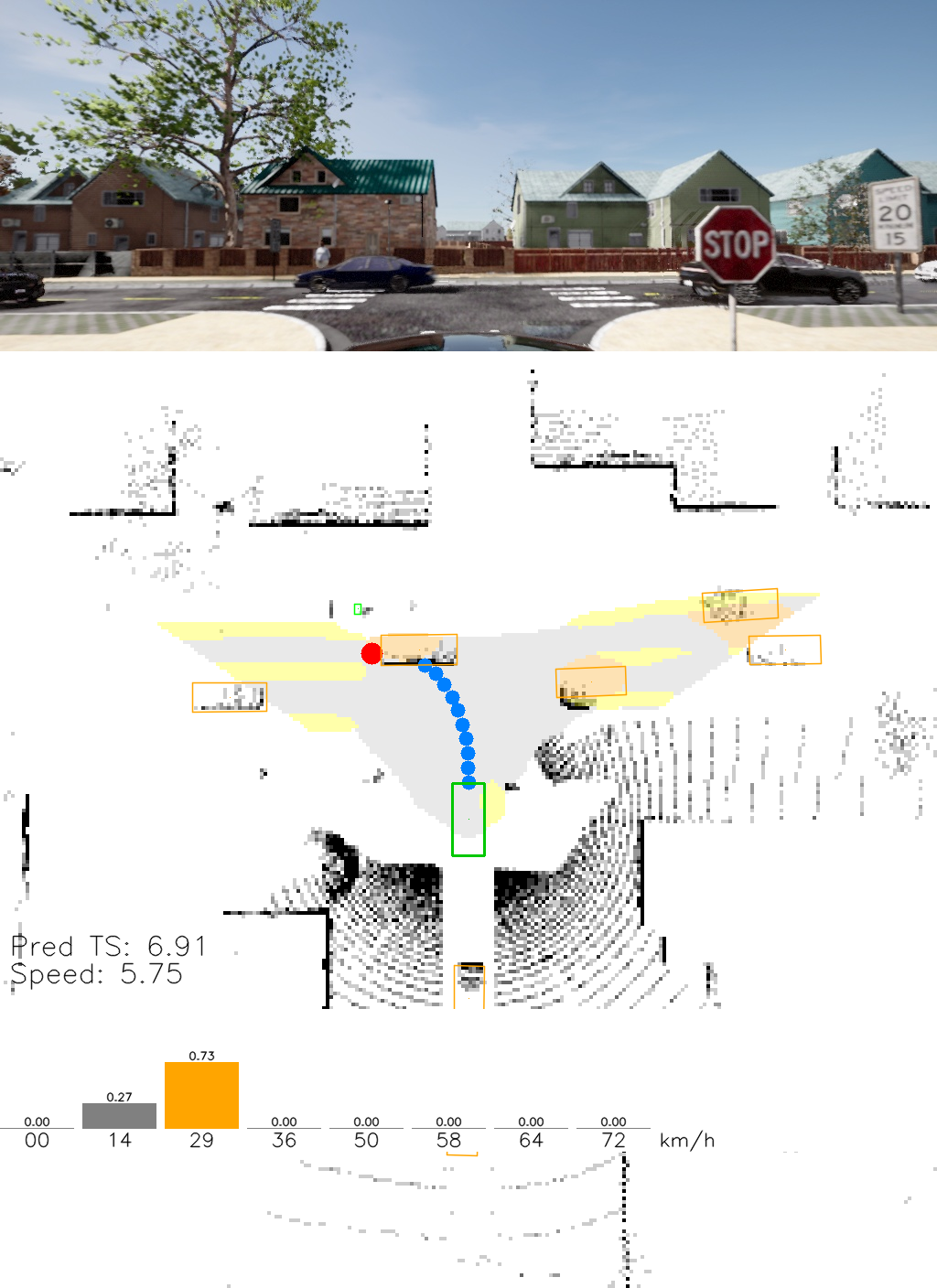}
    \hfill
    \includegraphics[width=0.328\textwidth, clip=true, trim=150pt 300pt 150pt 0pt]{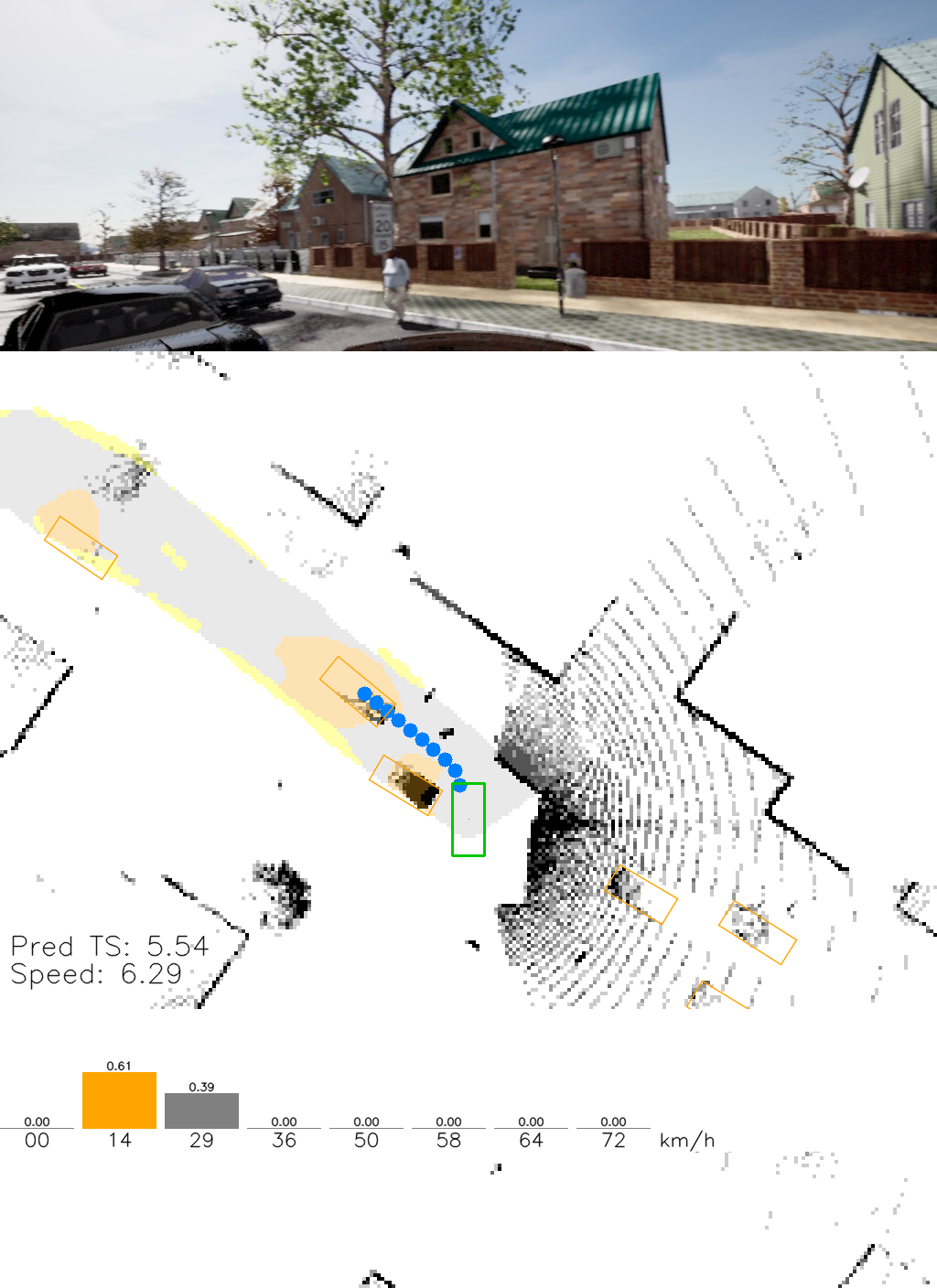}
    \hfill
    \includegraphics[width=0.328\textwidth, clip=true, trim=150pt 300pt 150pt 0pt]{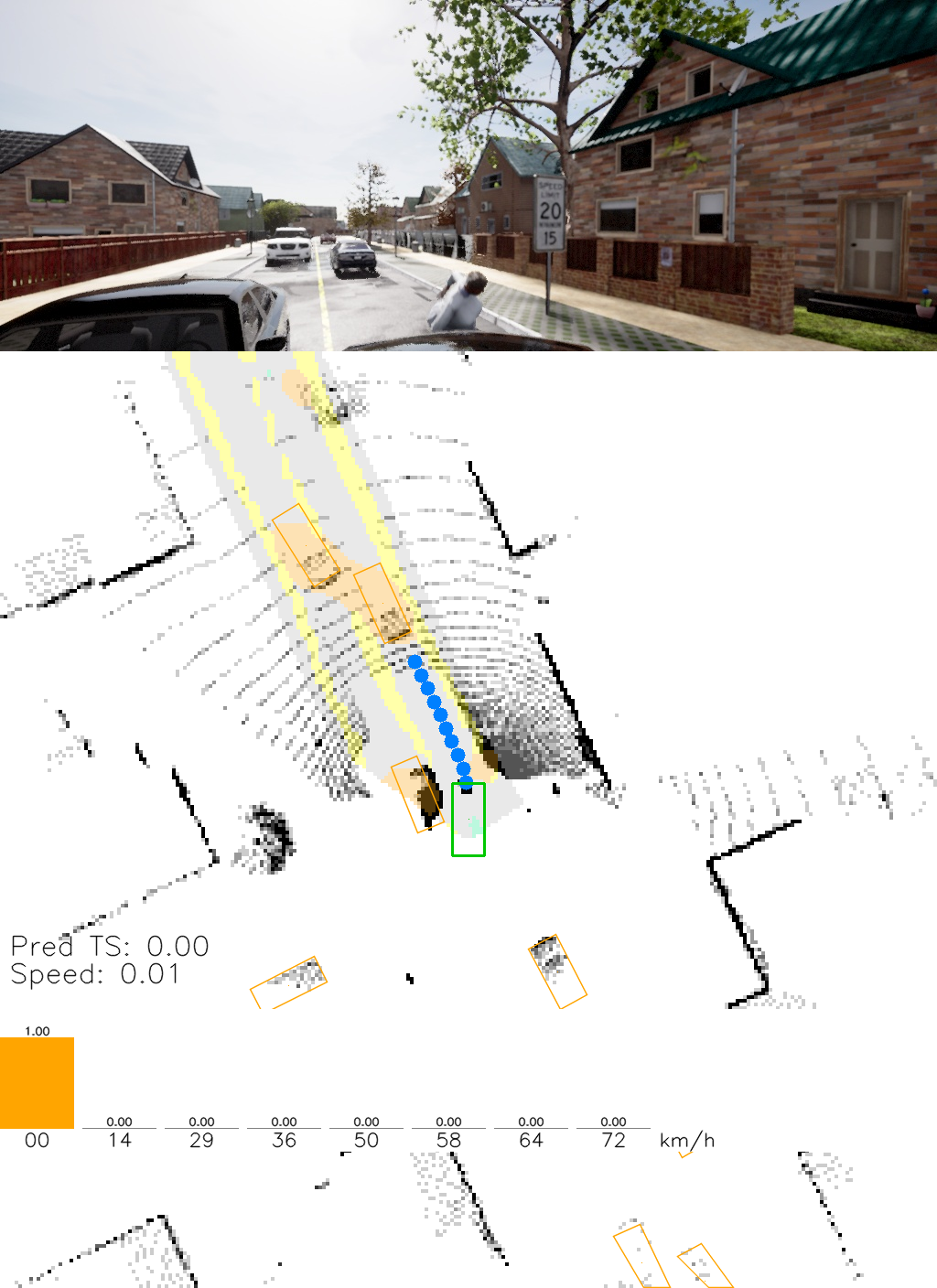}

    \vspace{-0.2cm}
    \rule{\textwidth}{0.4pt}
    \vspace{-0.2cm}

    \includegraphics[width=0.328\textwidth, clip=true, trim=150pt 300pt 150pt 0pt]{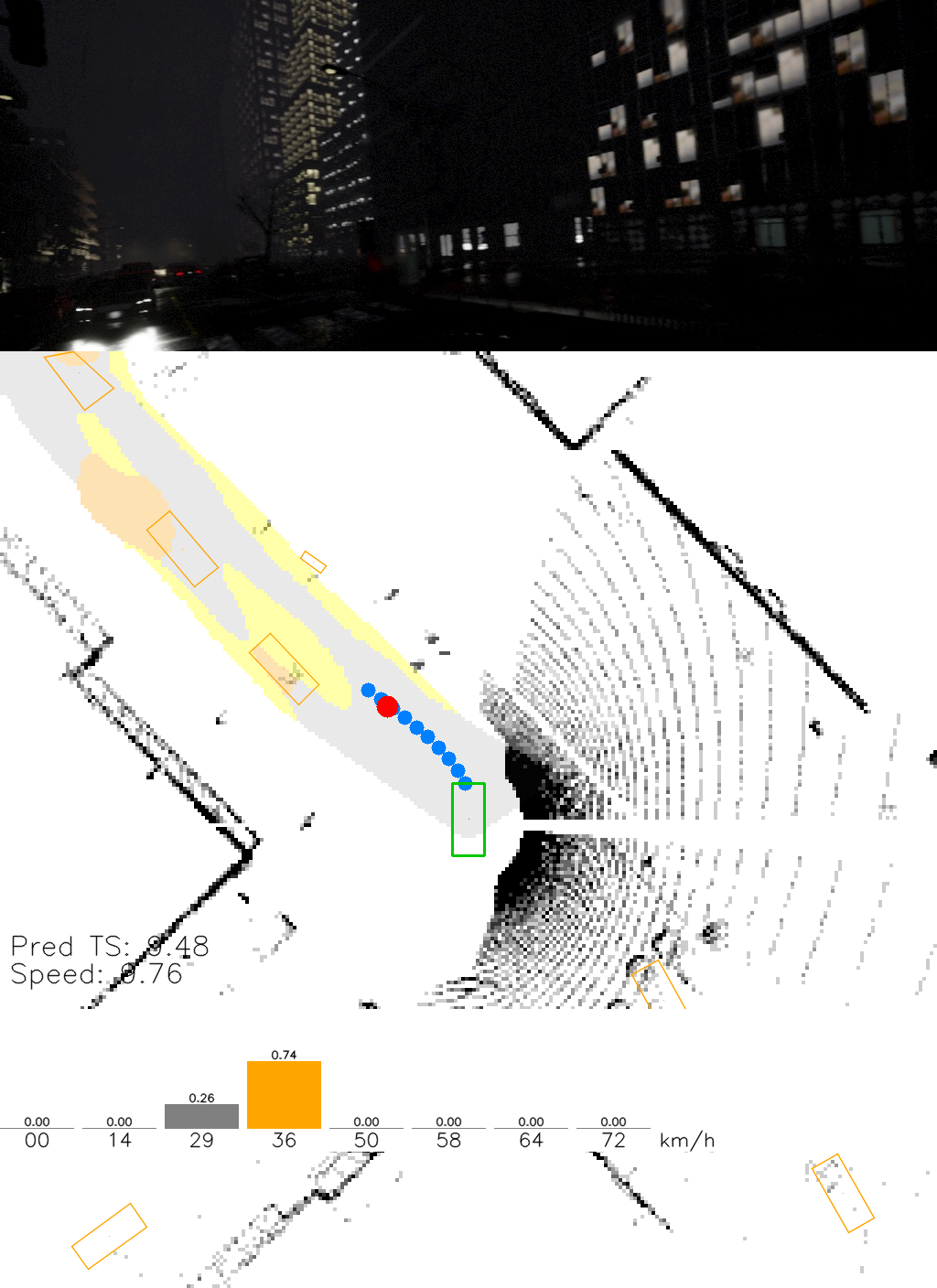}
    \hfill
    \includegraphics[width=0.328\textwidth, clip=true, trim=150pt 300pt 150pt 0pt]{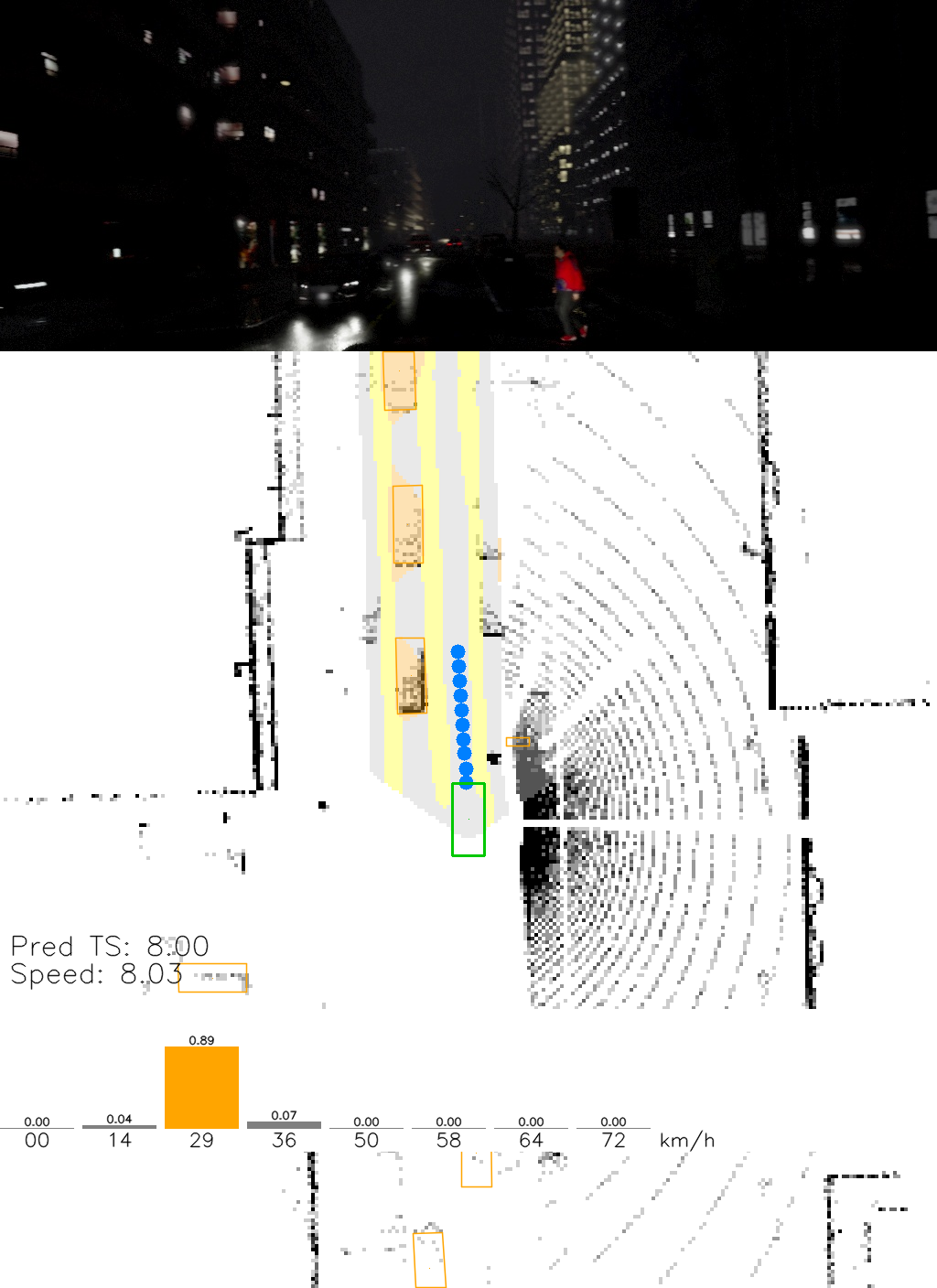}
    \hfill
    \includegraphics[width=0.328\textwidth, clip=true, trim=150pt 300pt 150pt 0pt]{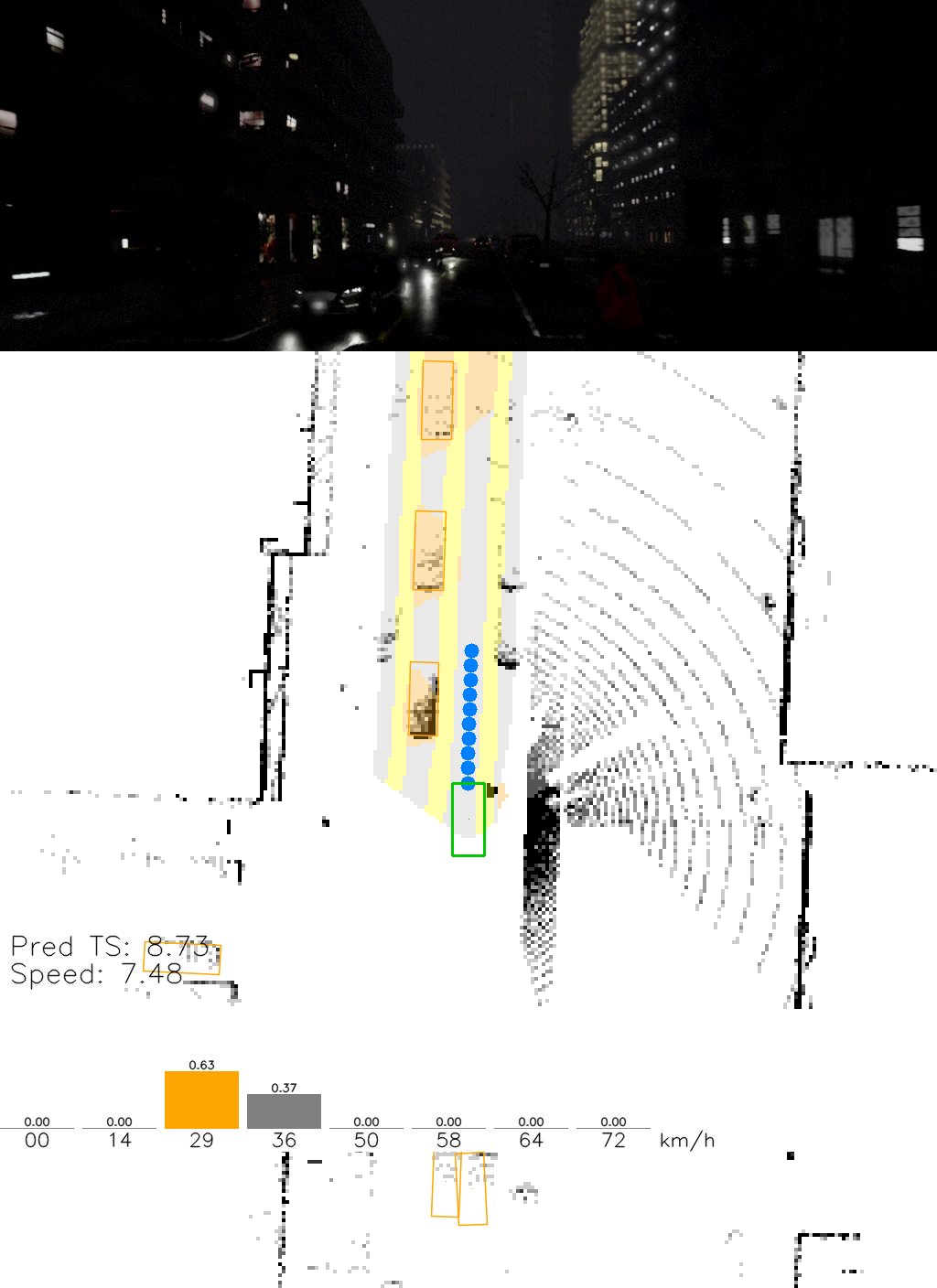}
\end{center}
\vspace{-0.6cm}
\caption{\textbf{VehicleTurningRoutePedestrian.} Top: Failure to perform unprotected turn. Bottom: Failure to recognize pedestrian at night.}
\label{fig:Scen_VTRP}
\vspace{-0.5cm}
\end{figure*}

%% file: sec_5_benchmarking.tex
\section{Benchmarking}
\label{sec:benchmarking}

Reliably benchmarking autonomous driving policies is challenging. While benchmarks based on real datasets exist, they are limited to open-loop metrics~\cite{Caesar2020CVPR,Dauner2024NEURIPS} or closed-loop evaluation of planning only, assuming privileged access to perception~\cite{Caesar2021CVPRW,Karnchanachari2024ICRA,Gulino2023NIPS,Chitta2024ECCV}. Therefore, simulators such as CARLA are essential for rigorous benchmarking of end-to-end driving policies. In the setting of CARLA Leaderboard 2.0, which we explore, there are two public benchmarks, on which we present our results in this section.

\subsection{Bench2Drive}

Bench2Drive~\cite{Jia2024NEURIPS} consists of 220 short ($\sim$150m) routes split across all CARLA towns, with one safety critical scenario in each route. It can be considered a `training' benchmark (reminiscent of level 4 autonomy), since methods under evaluation have seen the test towns during training.

\input{tab/tab_bench2drive}

\boldparagraph{Metrics} Besides the standard CARLA metric DS, this benchmark tracks success rate, which is 100\% for a given route if DS=100\%, and 0\% otherwise. Further, the 220 routes are categorized into five groups (Merging, Overtaking, Emergency Braking, Giving Way, and Traffic Signs), for which success rates are reported per category. 

\boldparagraph{Results} As seen in \tabref{tab:b2d}, TF++ significantly outperforms all prior work on this benchmark, doubling the success rate compared to the next best approach. We believe this is partly due to the higher performance of the PDM-Lite expert driver used in our dataset, compared to the closed-source Think2Drive~\cite{Li2024ECCV} expert used to generate training data for all the baselines in this setting.

\subsection{Official Town13 Validation Routes}

This is the set of 20 long validation routes on Town13 described earlier in \secref{sec:prelim}. As its name suggests, this is a `validation' benchmark, so data from Town 13 may not be used during training (reminiscent of level 5 autonomy).

\boldparagraph{Early termination} For this benchmark, it can be beneficial to stop an agent preemptively, reducing RC and increasing IS to improve DS. To illustrate this, we formulate the expected DS of an agent in terms of $x\in[0,1]$, which is the fraction of the route that the agent completes. Let $L$ be the route length, and $I = 0.5^{\frac{CP}{d}}*0.6^{\frac{CV}{d}}*0.65^{\frac{CL}{d}}*0.7^{\frac{RL}{d}}*0.7^{\frac{SI}{d}}*0.7^{\frac{ST}{d}}*0.8^{\frac{YE}{d}}$ be the expected infraction coefficient per km, including all non-negligible infraction types (collisions with pedestrians, collisions with vehicles, collisions with layout, red light infractions, stop infractions, scenario timeouts, and yield to emergency vehicle infractions). Since we divide the exponents of $I$ by the distance $d=xL$ traveled by the agent, the IS can be calculated as $I^{xL}$, and RC as $100x$, giving $DS=100xI^{xL}$. Maximizing this function, we obtain the solution 
$x_{max} = -(L\:\cdot\:\text{log}\,I)^{-1},$
with a theoretically maximal driving score of $DS(x_{max}) = -\frac{100}{L\: \cdot \: e \:\cdot\: \text{log}\,I}$. Figure~\ref{fig:expected_ds_curves} plots this function, along with its maxima, for different values of $I$.

\input{fig/fig_expected_ds_curves}

We observe that for benchmarks with long routes like Town13 validation and the official leaderboard, mathematically, a model profits from "early termination" if $x_{max} < 1$, i.e., $I<0.907$. The plot illustrates this: if $I<0.907$,  expected driving score is maximized at $x_{max} < 1$, with driving scores dropping off significantly at higher route completion fractions $x$. With $L=10.295$ and $I = 0.43$ (estimated for our model on the validation routes), we obtain $x = 0.115$. Thus, we should theoretically stop at $d = xL = 1.18\mathrm{km}$ to maximize the expected DS. Since $I$ and $L$ are only estimates, we set target speed to 0 after $d = 1.5\mathrm{km}$ in practice. We track distance traveled using the agent's speed sensor. As \tabref{tab:leaderboard} shows, all good submissions to the Leaderboard 2.0 test server have less than 18.1 RC, which implies that all these methods use a variant of early termination either explicitly or implicitly.

\boldparagraph{Normalized DS} This tradeoff introduced by the performance metrics used in Leaderboard 2.0 forces participants to terminate evaluations early to remain competitive, which is counterproductive. Therefore, we recommend adjusting the performance metrics for future challenges. Instead of using the infraction score ($IS$) for driving score calculation, we propose using the infraction coefficient ($I$) as defined above, which incorporates infraction \emph{frequencies}:
\vspace{-0.1cm}
$$\widehat{DS}(x) = RC(x) \times I(x) \approx 100xI, \:x\in[0,1].$$

This function, depicted in Figure~\ref{fig:expected_ds_curves_changed} for varying values of $I$, increases linearly with $x$. This eliminates the incentive to stop early, since the maximum driving score is always reached at $x=1$, i.e., full route completion. As a side effect, this change leads to an increase of average driving scores achieved with identical models compared to the original formula. If this increase is not desired, it can be corrected by scaling down the penalty factors for all infractions, which reduces the expected driving scores while maintaining the maximum score at 100. 

\boldparagraph{Results} We first demonstrate the impact of our metric adjustment on the official Town13 validation routes. Early termination has a significant effect on the scores obtained, as shown in Table \ref{tab:early_stopping}. Here, we scale all infraction penalties by a factor of 0.2 for $\widehat{DS}$ to keep the resulting normalized driving scores in a similar range as with the original formula. Concretely, we apply a base penalty of $0.2 * 0.5 = 0.1$ for $CP$, $0.2 * 0.6 = 0.12$ for $CV$, and so forth. Comparing this to the original metrics, it is clear that the revised driving score calculation successfully discourages early termination, since models that stop after one kilometer now receive a much lower normalized driving score, as intended. Comparing TF++ to UniAD\footnote{{publicly available reimplementation of UniAD-Base for CARLA, at \texttt{\url{https://github.com/Thinklab-SJTU/Bench2DriveZoo}}}}, we see large improvements in terms of both DS and $\widehat{DS}$. Finally, we note the significant gap between TF++ and the PDM-Lite expert, indicating that this is a promising benchmark for future research.

%% file: tab/tab_bench2drive.tex
\begin{table*}[t!]
\small
\centering
\begin{tabular}{l|cc|rrrrr|r}
\toprule
\multirow{2}{*}{\textbf{Method}} & \multicolumn{2}{c}{\textbf{Overall}} & \multicolumn{6}{|c}{\textbf{Multi-Ability}}  \\
\cmidrule{2-9} 
& Driving Score $\uparrow$  & Success Rate $\uparrow$ & Merge $\uparrow$  & Overtake $\uparrow$ & EmgBrake $\uparrow$  & GiveWay $\uparrow$ & TSign $\uparrow$  & Mean $\uparrow$ \\
\midrule
AD-MLP~\cite{zhai2023ADMLP} & 18.05 & 0.00 & 0.00 & 0.00 & 0.00 & 0.00 & 4.35 & 0.87 \\
TCP~\cite{Wu2022NeurIPS} & 40.70 & 15.00 & 16.18 & 20.00 & 20.00 & 10.00 & 6.99 & 14.63\\
VAD~\cite{Jiang2023ICCV} & 42.35 & 15.00 & 8.11 & 24.44 & 18.64 & 20.00 & 19.15 & 18.07\\
UniAD~\cite{hu2023_uniad} & 45.81 & 16.36 & 14.10 & 17.78 & 21.67 & 10.00 & 14.21 & 15.55 \\
ThinkTwice~\cite{Jia2023CVPR} & 62.44 & 31.23 & 27.38 & 18.42 & 35.82 & \textbf{50.00} & 54.23 & 37.17\\
DriveAdapter~\cite{Jia2023ICCV} & 64.22 & 33.08 & 28.82 & 26.38 & 48.76 & \textbf{50.00} & 56.43 & 42.08\\
TF++ (Ours) & \textbf{84.21} & \textbf{67.27} & \textbf{58.75} & \textbf{57.77} & \textbf{83.33} & 40.00 & \textbf{82.11} & \textbf{64.39} \\
\midrule
\textit{PDM-Lite}~\cite{Sima2024ECCV} & \textit{97.02} & \textit{92.27} & \textit{88.75} & \textit{93.33} & \textit{98.33} & \textit{90.00} & \textit{93.68} & \textit{92.82}\\
\bottomrule
\end{tabular}
\caption{\textbf{Results on Bench2Drive}. Note that the baselines use Think2Drive~\cite{Li2024ECCV} as an expert, while TF++ uses PDM-Lite.
}
\label{tab:b2d}
\vspace{-0.4cm}
\end{table*}

%% file: fig/fig_expected_ds_curves.tex
\begin{figure}[t!]
\begin{center}
    \includegraphics[width=\columnwidth]{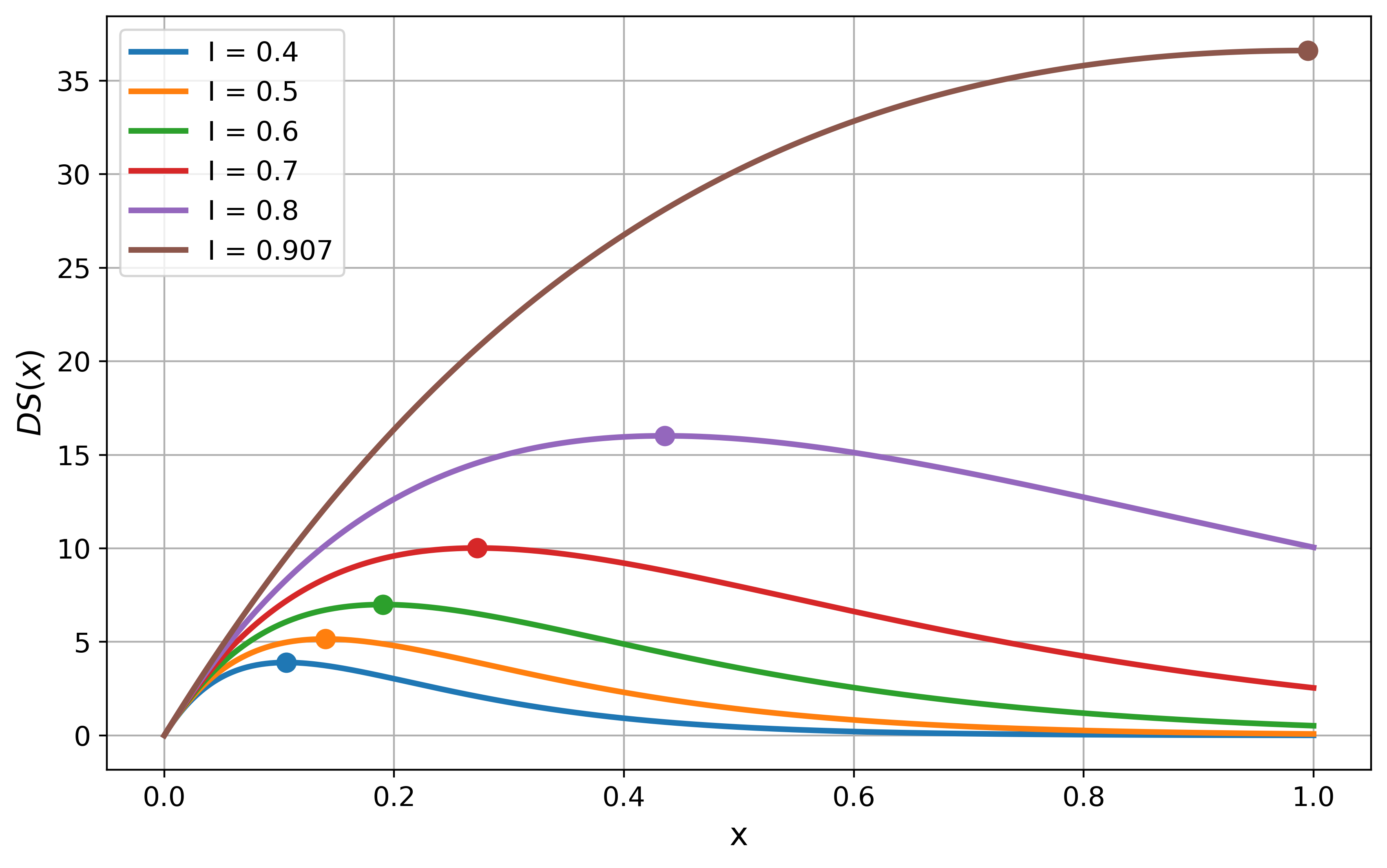}
\end{center}
\vspace{-0.7cm}
\caption{\textbf{Approximate DS as a function of RC fraction $x$ for different infraction coefficients $I$.} $DS(x)$ has a global maximum at $x<1$ if $I<0.907$, which creates an incentive to stop early.}
\label{fig:expected_ds_curves}
\vspace{-0.3cm}
\end{figure}

%% file: sec_6_conclusion.tex
\section{Conclusion}
\label{sec:counclusion}

With a systematic analysis of training dataset biases for end-to-end driving in CARLA, we reveal the impact of expert style on IL policy performance, provide insights into the challenges of assigning importance to frames through weighting or filtering, and provide a simple yet effective heuristic that estimates importance based on changes in target labels. We reproduce TransFuser++ in the Leaderboard 2.0 setting, providing the first recipe for training an end-to-end driving system that attains non-trivial performance. Finally, we propose an improvement to the existing metrics and extensively benchmark our system. We hope this can serve as a starting point for future research on this task.

\input{fig/fig_expected_ds_curves_changed}

\input{tab/tab_early_termination}

\boldparagraph{Limitations} In repeated Leaderboard submissions, we observe significant variance in DS, with identical agents yielding results that differ by more than 1 DS. Unfortunately, our results on the leaderboard (\tabref{tab:leaderboard}) also do not include all routes in the benchmark, due to technical issues during model setup in some routes. In addition, as shown in \tabref{tab:early_stopping}, DS is influenced significantly by early termination on long routes, which does not reflect any actual improvement in driving behavior. We believe it is necessary to consider improved metrics (such as the proposed Normalized DS) and standardize benchmarking with multiple seeds before drawing strong conclusions based on our empirical results.

%% file: fig/fig_expected_ds_curves_changed.tex
\begin{figure}[t!]
\begin{center}
    \includegraphics[width=\columnwidth]{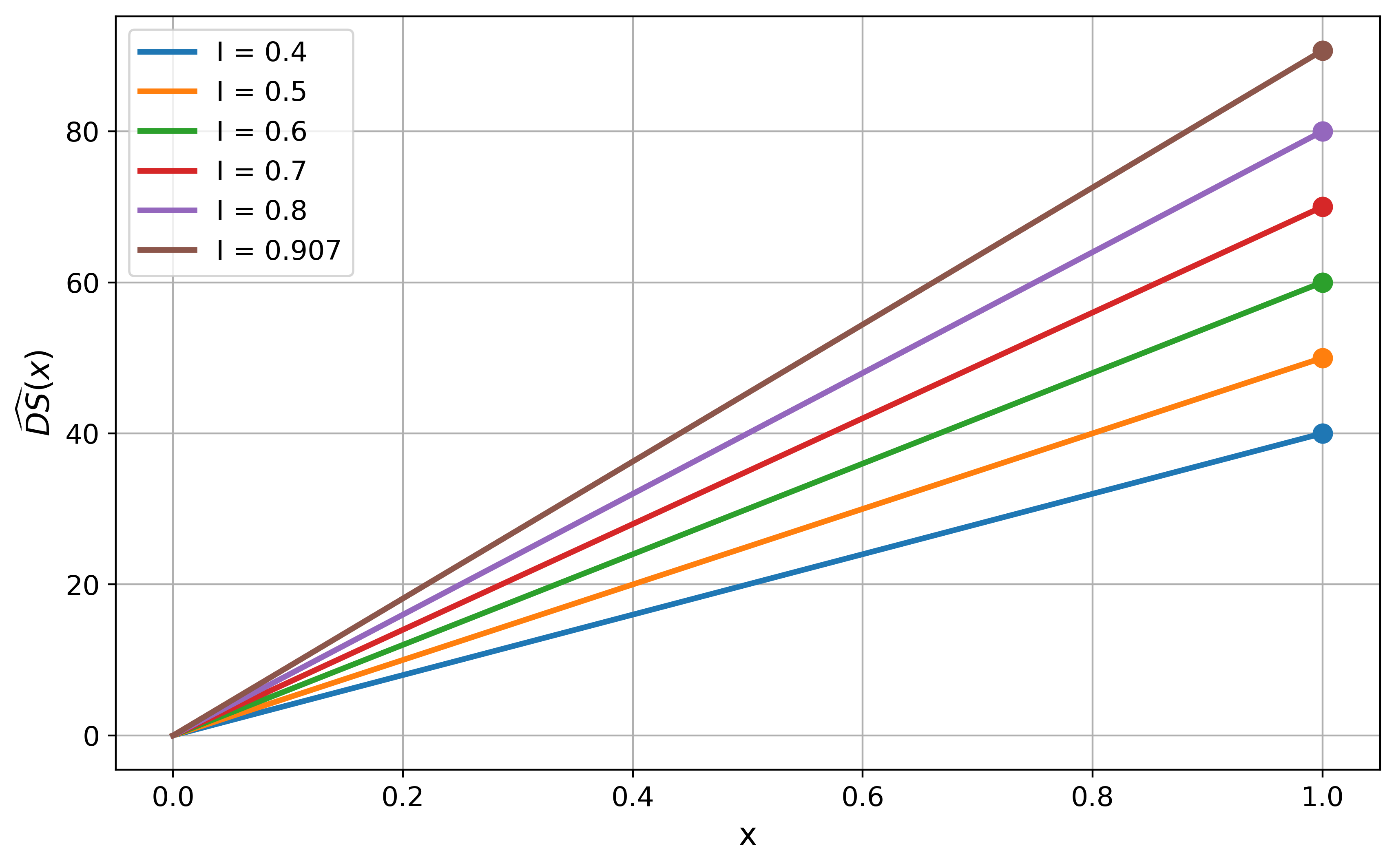}
\end{center}
\vspace{-0.7cm}
\caption{\textbf{Our proposed adjustment.} $\widehat{DS}(x)$ increases linearly with $x$, eliminating the incentive for early stopping.} %
\label{fig:expected_ds_curves_changed}
\vspace{-0.1cm}
\end{figure}

%% file: tab/tab_early_termination.tex
\begin{table}
\small
\setlength{\tabcolsep}{5pt}
    \centering
    \begin{tabular}{l | c | r | rr | rr}
        \toprule
        \textbf{Method} & \textbf{ET?} & ${\mathbf{RC}}$ $\uparrow$ & ${\mathbf{IS}}$ $\uparrow$ & ${\mathbf{DS}}$ $\uparrow$ & $\mathbf{I}$ $\uparrow$ & $\widehat{\mathbf{DS}}$ $\uparrow$ \\
        \midrule
        \multicolumn{7}{c}{{\color{lgray}{\textbf{Trained on all towns*}}}} \\
        \midrule
        {\color{lgray}{UniAD~\cite{hu2023_uniad}}} & {\color{lgray}{\xmark}} & {\color{lgray}{1.42}} & {\color{lgray}{0.49}} & {\color{lgray}{0.23}} & {\color{lgray}{0.30}} & {\color{lgray}{0.00}} \\
        \midrule
        \multirow{2}{*}{{\color{lgray}{TF++ (Ours)}}} & {\color{lgray}{\xmark}} & {\color{lgray}{68.53}} & {\color{lgray}{0.04}} & {\color{lgray}{0.96}} & {\color{lgray}{0.07}} & {\color{lgray}{4.94}} \\
        & {\color{lgray}{\cmark}} & {\color{lgray}{11.47}} & {\color{lgray}{0.43}} & {\color{lgray}{5.10}} & {\color{lgray}{0.18}} & {\color{lgray}{2.27}} \\
         \midrule
         \multicolumn{7}{c}{\textbf{Town13 withheld from training}} \\
        \midrule
         \multirow{2}{*}{TF++ (Ours)} & \xmark & 50.20 & 0.10 & 1.08 & 0.04 & 2.12 \\
         & \cmark & 10.91 & 0.34 & 3.73 & 0.12 & 1.30\\
        \midrule
        \textit{PDM-Lite}~\cite{Sima2024ECCV} & \xmark & 92.35 & 0.44 & 40.20 & 0.65 & 61.55 \\
        \bottomrule
    \end{tabular}
    \caption{\textbf{Results on official Town13 validation routes.} Mean over 3 evaluations of each agent. The normalized driving score $\widehat{\mathrm{DS}}$ removes the incentive for early termination ("ET?") after 1km on these long routes. {\color{lgray}{*We include this result for fair comparison to UniAD, however, we {strongly recommend} the setting with Town13 withheld for future work on this benchmark.}}}
    \label{tab:early_stopping}
    \vspace{-0.5cm}
\end{table}